\documentclass[11pt,,twocolumn]{article}

\usepackage[margin=1in]{geometry}
\usepackage[numbers]{natbib}
\usepackage{graphicx}
\usepackage{amsmath, amssymb}
\usepackage{setspace}
\usepackage{titlesec}
\usepackage{fancyhdr}
\usepackage{authblk}
\usepackage{hyperref}
\usepackage{booktabs}
\usepackage{multicol}
\usepackage{setspace}
\usepackage{titlesec}

\usepackage{multirow}
\usepackage{listings}
\DeclareUnicodeCharacter{03B5}{$\epsilon$} 
\usepackage{array}

\titleformat{\section}
  {\normalfont\Large\bfseries\centering} 
  {\thesection.}{1em}{}                  

\titleformat{\subsection}
  {\normalfont\large\bfseries\centering}
  {\thesubsection.}{1em}{}

\title{Predicting Anthropometric Body Composition Variables Using 3D Optical Imaging and Machine Learning} 

\author{Gyaneshwar Agrahari$^{*}$\textsuperscript{1}, Kiran
Bist \textsuperscript{1}, Monika Pandey \textsuperscript{1}, Jacob Kapita \textsuperscript{1}, Zachary James \textsuperscript{1},\\
Jackson Knox \textsuperscript{1},
Steven Heymsfield \textsuperscript{2}, Sophia Ramirez \textsuperscript{2}, Peter Wolenski \textsuperscript{1}, Nadejda Drenska \textsuperscript{1}}

\date{\footnotesize $^{*}$Corresponding author: gyan.agrahari77@gmail.com \\
\textsuperscript{1}Louisiana State University\\ \textsuperscript{\textbf{2}}Pennington Biomedical Research Center, Louisiana State University
}

\begin{document}

\maketitle

\begin{abstract}
    
{\small\itshape
  Accurate prediction of anthropometric body composition variables, such as Appendicular Lean Mass (ALM), Body Fat Percentage (BFP), and Bone Mineral Density (BMD), is essential for early diagnosis of several chronic diseases. Currently, researchers rely on Dual-Energy X-ray Absorptiometry (DXA) scans to measure these metrics; however, DXA scans are costly and time-consuming.
   This work proposes an alternative to DXA
scans by applying statistical and machine learning models on biomarkers (height, volume, left calf circumference, etc) obtained from 3D optical images. 
The dataset consists of 847 patients and was sourced from Pennington Biomedical Research Center.
Extracting patients' data in healthcare faces many technical challenges and legal restrictions. However, most supervised machine learning algorithms are inherently data-intensive, requiring a large amount of training data.
To overcome these limitations,
we implemented a semi-supervised model, the $p$-Laplacian regression model. This paper is the first to demonstrate the application of a $p$-Laplacian model for regression.
Our $p$-Laplacian model yielded errors of $\sim13\%$ for ALM, $\sim10\%$ for BMD, and $\sim20\%$ for BFP when the training data accounted for 10 percent of all data.
Among the supervised algorithms we implemented, Support Vector Regression (SVR) performed the best for ALM and BMD, yielding errors of  $\sim 8\%$ for both, while Least Squares SVR performed the best for BFP with $\sim 11\%$ error when trained on 80 percent the data.
Our findings position the $p$-Laplacian model as a promising tool for healthcare applications, particularly in a data-constrained environment. 
\par}
\end{abstract}

\section{Introduction}
\label{sec:introduction}

With the rise of applications in artificial intelligence in healthcare, personalized treatment and medicine have seen tremendous advances \cite{bohr2020rise}.
Estimating biometrics  accurately is essential to gaining insights into personalized human health.
Health researchers and nutritionists use several types of biometrics- metabolic (cholesterol, insulin), performance-based (speed, balance) and physiological (lean mass, body fat).
Appendicular Lean Mass (ALM), Body Fat Percentage (BFP), and Bone Mineral Density (BMD) are three of the most significant physiological biometrics.
ALM is linked to malnutrition risk and age-related frailty, and serves as a risk factor for adverse treatment outcomes associated with metabolic syndrome (MetS) and other clinical conditions \cite{bennett2024trunk}. 
BFP is associated with obesity and cardiovascular diseases \cite{Zeng_BFP}.
Similarly, BMD is instrumental to determining osteoporosis condition \cite{Fathima_BMD}.

Researchers have been using DXA or Dual-energy X-ray Absorptiometry to estimate these biometrics directly. Using a scan like DXA, however, is costly, time-consuming, and inconvenient for both researchers and patients.
An alternative approach is to estimate the biometrics by applying mathematical models to body measurements like height, weight, leg volume, and waist circumference.
This approach involves two key steps: (1) extraction of the body measurements and (2) applying a mathematical model on these measurements.
There have been developments in both these two key steps. 
Recent advancements now enable us to capture 2D and 3D digital images of people not only at research labs and hospitals but also at home using smartphone applications \cite{mccarthy2023}.
The body measurements can be extracted from these images using a computer program.
Such advancements have made the extraction of the body measurements cost-effective and time-efficient. 
For the second key step, traditionally, statistical tools like
 Anthropometric body composition prediction equations have been used to make the estimations, which typically do not account for non-linear relationships \cite{marin2022criterion}. In the past decade, however, machine learning algorithms have been proven to be more accurate and more efficient for estimating biometrics \cite{higgins2021machine}. 
Models, especially the supervised algorithms, require a substantial amount of data to be trained on \cite{adadi2021dataefficient}. 
However, the training data is typically acquired through imaging techniques such as DXA, which are costly to procure and operate.

In recent years, several machine learning algorithms have been developed that require less data to be trained on without significant loss of accuracy \cite{wang2020fewshot}.
Specifically, semi-supervised algorithms, such as
Semi-supervised support vector machines and graph-based label propagation have been proven efficient when training data is scarce \cite{semisupervised-survey}. 
In recent years, researchers used semi-supervised algorithm in many bioinformatics applications including predict anthropometric body compositions. 
Azadifar and Ahmdadi \cite{azadifar2022} used a semi-supervised learning method based on graph convolutional networks to classify and rank candidate disease genes.
Their model outperformed many similar models when tested on sixteen diseases \cite{azadifar2022}.
Zheng et.al\cite{Zheng_2021} used a semi-supervised learning self-training algorithm to predict BMD plain hip X-ray images BMD estimation method and achieved high Pearson correlation coefficient of 0.8805.

In this paper, we will focus on such a semi-supervised algorithm - $p$-Laplacian based regression.
We will use it alongside several supervised learning algorithms to predict the three biometrics: ALM, BFP, and BMD.




The paper is structured as follows. In Section \ref{sec:Literature Review}, we highlight the existing literature on predicting ALM, BFP and BMD using statistical and machine learning algorithms.
In Section \ref{sec:Data-Extraction}, we briefly describe how the anthropometric measurements were obtained. Section \ref{sec:data-description} includes a brief data summary. In Section \ref{sec:Methodology}, we explain the methodology of our analysis. Section \ref{sec:supervised-alg} briefly describes the supervised algorithms we implemented, whereas Section \ref{sec:semi-supervised-alg} does that for only semi-supervised algorithm - $p$-Laplacian based regression - in more detail. Section \ref{sec:results} encompasses all the main results from our analysis. We discuss our main findings and future work in Section \ref{sec:discussion}.

\section{Literature Review}
\label{sec:Literature Review}
Previous studies such as \cite{Wu_BMD, Shioji_BMD, AlvesML} have shown that machine learning models are accurate in predicting ALM, BMD, or BFP. Wu, \textit{et al}. in \cite{Wu_BMD}, used random forest, gradient boosting, neural network, and
 linear regression models to predict BMD.
They found that the gradient boosting model performed the best on a dataset comprised of $1103$ individual Single Nucleotide Polymorphisms (SNPs).
However, when they used Genetic Risk Scores (GRS's) as input data, all models performed similarly to one another.

Shioji \textit{et al.} in \cite{Shioji_BMD} used a dataset of 135 Japanese women of ages over $50$ to train and predict BMD and bone loss rates using artificial neural networks \cite{Shioji_BMD}. The authors discovered that neural networks performed better than multiple regression approaches.

Alves, \textit{et al.} in \cite{AlvesML} discovered that the gender of a person influences the prediction of BFP.
They used a dataset composed of both male ($84$) and female ($79$) subjects and developed various machine learning models to predict BFP on the combined dataset as well as on the datasets split up by gender.
The model types used to predict BFP were random forest regression, extreme gradient boosting,
 decision tree, support vector regression, multi-layer perceptron (MLP) regression,
 and least square support vector regression (LSSVR).
They observed that when using the combined dataset, the mean absolute error (MAE) range
 was between $3.569$ and $9.859$;
 the LSSVR achieved the best performance, and MLP reached the worst when using the combined dataset \cite{AlvesML}.
However, the BFP error range decreased to a low of $2.756$ with LSSVR and a high of $6.15$ with MLP
 when the dataset was restricted to males only.
For the females, the error ranged from $4.004$ with LSSVR
 to $8.003$ with MLP.

Marazzato, \textit{et al.} \cite{Marazzato_alm} used a neural network with $1$ input, $3$ hidden, $2$ dropout, and $1$ output layers
 to predict ALM from $10$ demographic and $43$ digital anthropometric measurements aquired using a 3D optical scanner \cite{Marazzato_alm}.
The used a dataset with $576$ subjects.
The results, showing small mean, absolute, and root-mean square errors, 
 helped Marazzato \textit{et al.} to recommend the use of neural networks to predict ALM.

Barbosa \textit{et al.} used a dataset of $190$ participants ($118$ women and $72$ men) to predict ALM using muscle thickness measurements extracted via ultrasound \cite{Barbosa_ALM}.
They recommended using two prediction equations after observing that the developed equations produced unbiased ALM estimates that were close to the reference measurements found by the Ultrasound scanner.

There have been limited applications of semi-supervised (SSL) algorithms in predicting ALM, BMD, and BFP.
Zheng et.al\cite{Zheng_2021} used a semi-supervised learning self-training algorithm to predict BMD plain hip X-ray images BMD estimation method, and achieved a high Pearson correlation coefficient of 0.8805.
Our work initiates the application of SSL algorithms in the prediction of anthropometric body composition variables. 
To our knowledge, this technique has not been used to predict ALM, BMD, or BFP in previous research.

\section{Data Extraction}
\label{sec:Data-Extraction}
 All the body measurements, here referred to as biomarkers, were gathered at the Pennigton Biomedical Research Center (PBRC) facility \cite{Sobhiyeh2021}. There were 847 
 participants, and each participant had measurements of biomarkers such as the circumference and/or length of chest, waist, hip, arm, thigh, etc. deduced from the 3D image. The image was created from 2D images 
 taken by one of the 3 different machines, namely (i.) the Proscanner, which uses three stationary cameras aligned vertically on a column to image a person standing on a rotating turntable, (ii.) the Styku machine, which has a similar design as the Proscanner but with a single camera, and finally (iii.) the SS20, which has 5 cameras positioned on 4 different vertical columns to capture the image \cite{Sobhiyeh2021}.  

The images from the machines were used to create a 3D triangular mesh with the 3D vertices representing cloud points and the area between them making a surface. Then, software created in Matlab by LSU's Math Consultation Clinic ($MC^2$) group corrects any errors in the meshing \cite{Sobhiyeh2021}, such as any errors in the mesh, correcting surfaces and filling any holes in the mesh. 

After the mesh is corrected, it is segmented into various parts of the body which are the center, right arm, right leg, left arm, and left leg. Then each data point is taken from the segmentation which creates the dataset of body parts from each patient \cite{Sobhiyeh2021}. Out of all the measured biomarkers, we have used 44 biomakers in this study.

\section{Data Description}
\label{sec:data-description}
Pennington Biomedical Research Center in Baton Rouge, Louisiana provided the data used for this research. The collected data was utilized to predict ALM (Appendicular Lean Mass), BMD (Bone Mineral Density), and BFP (Body Fat Percentage). Additionally, influenced by the dependence of the three target values on sex, we conducted an in-depth data analysis by dividing it into three datasets: Male, Female, and Combined (the entire dataset). Using a correlation matrix, we examined the correlation of the biomarkers with the target variables (ALM, BMD, and BFP) and identified the top 10 biomarkers for each dataset.

\subsection{Dataset Characteristics and Structure}
\label{subsec:data-character}
The dataset contains numerical and categorical data and is stored in a widely used CSV format.
The dataset’s structure includes rows representing individual participants and columns representing various
measurements. Each row contains the
data for a single participant. Similarly, the columns
include categorical variables like gender and race and
numerical variables like age, ALM, BMD, BFP, and
various body measurements.

\subsection {Data Cleaning}
\label{subsec:data-clean}
We removed participants with missing values in order to clean the data. 
After eliminating the rows with missing entries, we used the remaining clean data for the training and testing purposes.
The cleaned data is composed of a total of $515$ people and their respective $44$ biomarkers. During training and testing, we excluded some of columns such as "Site", "Race"  and "Gender". Finally the dataset is composed of 3 target variables and $44$ biomarkers.

\subsection {Descriptive Statistics}
\label{subsec:data-stat}
The dataset consists of 245 males and 270 females, showing a slightly higher number of female participants. In the combined dataset, the top ten biomarkers demonstrate a strong positive correlation with ALM (over 0.9), a moderate correlation with BMD (0.7 to 0.81), and a lower correlation with BFP (0.4 to 0.6). When analyzing the data separately, most of the biomarkers showed a stronger correlation with ALM and BMD in males than in females. However, among the top ten biomarkers, Surface Area Total shows the highest correlation with ALM in both sexes.


For BFP, the correlation with the top ten biomarkers
is slightly higher in females (0.6 to 0.76). In
both genders, Horizontal Waists showed the strongest
correlation with BFP. This high correlation is consistent with the research finding that adult women generally
have higher body fat percentages than men\cite{jackson2002effect}.

For BMD, the Surface Area Arm is most strongly associated with males, while Height is the most influential factor in females. Biologically, testosterone in males 
 contributes to greater BMD, while estrogen plays a crucial role in maintaining bone density in females \cite{vaananen1996estrogen}. However, after menopause, a significant drop in estrogen levels can lead to an increased risk of bone loss and osteoporosis in women.
 \begin{table}[h!]
    \centering
\begin{tabular}{c|lrrr}
\toprule
\multirow{7}{*}{\rotatebox{90}{\textbf{Male}}} 
 & {} & \textbf{Mean} &\textbf{ Median }& \textbf{SD }\\
\midrule
 & ALM (kg)           &  21.88 &   23.40 & 7.95 \\
 & BFP (\%)           &  25.20 &   24.96 & 7.67 \\
 & BMD           &   1.07 &    1.10 & 0.19 \\
 & Age           &  27.33 &   17.00 & 19.59 \\
 & Height (cm)   & 164.33 &  171.20 & 17.80 \\
 & Weight (kg)   &  68.24 &   69.70 & 25.88 \\
\midrule
\end{tabular}

    \begin{tabular}{c|lrrr}
\multirow{7}{*}{\rotatebox{90}{\textbf{Female}}} 
&ALM (kg)          &  15.77 &   15.50 &                4.24 \\
&BFP (\%) &  34.71 &   34.90 &                7.37 \\
&BMD   &   1.00 &    1.01 &                0.14 \\
&Age           &  27.88 &   17.00 &               20.25 \\
&Height (cm)   & 156.68 &  158.75 &              12.47 \\
&Weight (kg)   &  60.00 &   58.85 &               18.22 \\
\midrule
\end{tabular}
\begin{tabular}{c|lrrr}
\multirow{7}{*}{\rotatebox{90}{\textbf{Combined}}} 
&ALM (kg) & 18.68 & 17.90 & 6.98 \\
&BFP (\%) & 30.19 & 30.14 & 8.89 \\
&BMD& 1.03 & 1.04 & 0.17 \\

&Age & 27.62 & 17.00 & 19.92 \\
&Height (cm) & 160.32 & 162.40 & 15.70 \\
&Weight (kg) & 63.92 & 63.20 & 22.55 \\
\bottomrule
\end{tabular}
 

 \caption{Summary of Statistics Related to Male, Female, and Combined} 
    \label{tab:my_label}
\end{table}

Table 1 summarizes statistics for various physiological and demographic parameters across male, female, and combined datasets. Males display an average ALM of 21.88 kg and a BFP of 25.20\%, while females show lower ALM (15.77 kg) but higher BFP (34.71\%), with the combined data reflecting intermediate values (ALM at 18.68 kg, BFP at 30.19\%). Although ALM is slightly positively skewed, BFP and BMD distributions are balanced across all datasets. The data predominantly consists of younger individuals with a median age of 17, although older individuals raise the mean age to 27.62. Both height and weight show considerable variability, with high variance in the top ten biomarkers for BMD, BFP, and ALM. This variability also indicates diverse physiological measurements and notable individual differences within the sample.







‌ 

‌





\section{Methodology}
\label{sec:Methodology}
We tested all the models on three datasets - Male, Female, and Combined - for each target variable: ALM, BMD, and BFP. 
 We investigated the effects of normalizing the data using both the \texttt{StandardScaler()} and \texttt{MinMaxScaler()} classes from scikit-learn. The  \texttt{StandardScaler()} outperformed \texttt{MinMaxScaler()} in our experiments, so we use the \texttt{StandardScaler()} throughout this article.
To validate the models' analysis, 
we used $K$-Fold Cross Validation. For the semi-supervised model, we further modified the $K$-Fold cross validation to train the model on a smaller percentage of training data. In any case, we independently ran the models 10 times where in each run the data was split into $K$ folds randomly.
\subsection{K-Fold Cross Validation}
\label{subsec:method-k-fold}
$K$-fold cross-validation is a statistical method used to evaluate the performance and robustness of machine learning models. In $K$-fold cross-validation, the dataset is divided into $K$ equally sized subsets or folds. The training is performed on the $K-1$ folds and then tested on the remaining fold \cite{ James23, James}. The process of training and testing is then repeated $K$ times, so that each fold is used for testing in turns. Then, the $K$-fold results are averaged to produce a single performance metric. This method is necessary because it provides a more accurate estimate of a model's performance compared to a single train-test split, because it reduces the variance associated with the random division of data. Additionally, $K$-fold cross-validation helps identify whether a model is overfitting or underfitting by making sure that every data point is used for training and validation. Thus, it improves the generalizability of the model to unseen data and aids in hyperparameter tuning and model selection \cite{kcrossvalidation}.
\\


\subsection{Error Analysis}
\label{subsec:method-error}
We used Root Mean Square Error (RMSE) in percent, given by  Equation \ref{eqn:RMSE}, to optimize our models. Our goal is to predict ALM, BFP, and BMD with the highest accuracy possible.  The RMSE is expressed as a percentage for better comparison across models, target variables, and datasets.

\begin{equation}
\label{eqn:RMSE}
        \text{RMSE}=\frac{\sqrt{\frac{1}{N}\sum_{i=1}^N(\hat{y}_i-y_i)^2}}{\sqrt{\frac{1}{N}\sum_{i=1}^N y_i^2}}\times 100
    \end{equation}
where,\\
$y_i=$ True value of ALM, BFP, or BMD,\\
$\hat{y}_i=$ Predicted value of ALM, BFP, or BMD, and\\
$N=$ Number of patients in the testing set.

Each model was implemented 10 times where each run was implemented with $K$-cross validation. The average RMSE for the model was calculated by the taking average over all runs and $K$-folds. For example, if $K=5$, the average was taken over 50.

\section{Supervised Learning Algorithms}
\label{sec:supervised-alg}
In this section, we present an overview of the supervised algorithms we implemented. For these models, we use the standard 80-20 split with $K$-fold cross-validation for $K=5$, hence hereon we will refer to it as $5$-fold cross validation. That is, $20\%$ of the data is used for testing in each fold.
In this section, we briefly describe the supervised models we use.
We discuss our results for these models in the Results and Analysis section.

\subsection{Linear and Polynomial Regression Models}
\label{subsec:supervised-regression}
Linear Regression (LR) and Polynomial Regression (PR) are widely used techniques in machine learning. However, they are susceptible to underfitting or overfitting and sensitive to outliers \cite{James}.
To improve their efficiency, researchers have developed variations of LR and PR \cite{variations-of-LR-PR}. In this study, we implemented three such variations—Lasso, Ridge, and Bayesian—along with traditional LR and PR. Due to time complexity, we implemented PR of degree 2 and degree 3 only.


\subsection{Support Vector Regression Models}
\label{subsec:supervised-svr}
Support Vector Regression (SVR) is an extension of regression models to higher dimensions with additional features. SVR applies a kernel function that maps the raw data into higher dimensions. The model then tries to fit the transformed data within an $\epsilon$ margin of a hyperplane. There are several choices for the kernel function \cite{SVR}.
In our experiments, the Linear kernel outperforms other kernel functions for SVR.
The penalty parameter $C$ allows us to choose the trade-off between the error and over-fitting.

We applied a variant of SVR called Least Square SVR (LSSVR) with a radial basis function (RBF) kernel function. LSSVR  solves a set of linear equations instead of quadratic programming like in SVR \cite{lssvr}.
The parameter $\gamma$  in LSSVR decides how influential one data point is. If $\gamma$ is low, the kernel output is close to 1 for most data points, leading to under-fitting.

For SVR, we varied the values of $\epsilon$ and $C$ whereas for LSSVR, we varied the values of $C$ and $\gamma$.


\subsection{Random Forest and Extreme Gradient Boosting Models}
\label{subsec:supervised-rf-xg}
The Random Forest (RF) model is an ensemble learning method that averages predictions from multiple decision trees for classification and regression problems. The decision trees are trained independently of each other on random subsets of the data \cite{Random-Forest}.
In the Extreme Gradient Boosting (XGBoost) model, the decision trees are added sequentially, where each new tree corrects the residual error of its predecessor. The model uses gradient descent to minimize loss. 
 We fine-tuned the number of trees and maximum depth of a tree for both RF and XGBoost to determine which combination of these parameters minimizes RMSE \cite{xgboost}.


\subsection{Neural Network (NN)}
\label{subsec:supervised-nn}
Neural networks mimic the way neurons in the human brain function. They are designed to model and understand patterns in data by adjusting weights and biases through training \cite{Goodfellow-et-al-2016}. The effectiveness of the model is indicated by a loss function. There are various types of neural networks suited for specific tasks. In this study, we focus on regression and have chosen a multi-layer perceptron (MLP). For the MLP, we experimented with up to four hidden layers, each associated with an activation function. The most optimal batch size was 16.
We selected ReLU as the activation function for all layers.
Among the optimizers tested (Adam, AdamW, RMSEprop, and
SGD), Adam yielded the best performance. MSE (Mean Squared Error) and RMSE (Root Mean Squared Error) were the loss functions used in the experiments. 


\section{Semi-Supervised Learning Algorithms}
\label{sec:semi-supervised-alg}
Semi-supervised learning is a category of machine learning algorithms that combines the utilities of unsupervised learning and supervised learning algorithms.
Supervised algorithms construct a classifier or a regressor by training the dataset to find associations between the input values and the target. 
Such algorithms then leverage these associations to predict the output value for previously unseen inputs \cite{James23}.
In an unsupervised algorithm, inputs are assigned to clusters based on feature similarity, ensuring that inputs with similar characteristics are grouped. In a semi-supervised algorithm, the relationship between labeled (training) and unlabeled (testing) data is utilized to improve the learning process \cite{James23}.
A semi-supervised algorithm is useful when extracting labeled data is either computationally or financially expensive.
Semi-supervised algorithms have traditionally been applied to classification tasks, as their design is often tailored specifically for classification and does not readily extend to regression problems.
The class of semi-supervised learning methods known as graph-based methods can be extended to regression problems \cite{semisupervised-survey}. Zhu \cite{Zhuthesis} is one of the first researchers to introduce graph-based semi-supervised algorithms in their thesis.
We implemented a version of a graph-based method, called $p$-Laplacian regression introduced by Calder and Drenska in 
\cite{Calder-Drenska}.
In this section, we will elucidate the underlying concepts of their model.
For the sake of simplicity, we will explain the model for predicting ALM of the patients. The same procedure works for BFP and BMD.
In the following discussion, by labeled vertex, we refer to a patient whose ALM value the algorithm remembers. Such vertices are in the training set for the algorithm. 

We discuss this graph-based method from \cite{Calder-Drenska} in two steps: (1) Graph Construction and (2) The Tug-of-War equation.

\subsection{Graph Construction}

We follow the standard notation and terminology for the graph. Let $G$ be our graph with $n$ vertices where $n$ is the total number of patients in the dataset.
Let $\mathcal{X}=\{x_1,x_2,\dots,x_n\}$ be the set of vectors in $\mathbb{R}^{44}$. Each $x_i$ represents a vertex in $G$. Each entry in $x_i$ is a measurement of one of the biomarkers of the patient the vertex represents.
The $x_i's$ are first normalized using $z$-score normalizer. 
We connect two vertices $x$ and $y$  with an edge only when they are within the threshold of the similarity we define below.
For a pair of vertices $x$ and $y$, we define the Euclidean distance, $d(x,y) =||x-y||$ in $\mathbb{R}^{44}$.
Intuitively, for the pair of vertices $x$ and $y$, the greater the value of $d(x,y)$ is, the less similar the patients corresponding to $x$ and $y$ are. 
Therefore,  we need a nonnegative decreasing function $\eta$ to define the edge weight, $w_{xy}$.
\begin{equation}
    \label{eqn:weight}
    w_{xy}=\eta\Bigg(\frac{d(x,y)}{\epsilon}\Bigg)
\end{equation}

where $\epsilon>0$ is a free parameter.

Equation \ref{eqn:weight} implies $w_{xy}=w_{yx}$. Therefore, $G$ is an undirected graph. 
As used in \cite{Calder-github},  we chose $\eta(t) =e^{-t^2}$ as our nonnegative decreasing function. The nondecreasing function allows us to connect nodes with nodes.
Furthermore, this choice ensures the values of $w_{xy}$ lie strictly between 0 and 1. 
For computational efficiency, we construct a $k$-nearest neighbors ($k$-NN) graph, where each vertex has $k$ neighbors.
We follow the method described by Calder and Trillos in \cite{Calder-Trillos} to construct a $k$-NN graph.
We denote the open neighborhood of a vertex $x$ as $N_x$.
Theoretically, this algorithm of constructing a $k$-NN works even when $G$ is not connected, provided that each connected component of $G$ has at least one labeled vertex. However, for the computer simulations to run, $G $ must be connected. 
The primary hypothesis underlying the construction of a similarity-based graph is that individuals with similar measurements of their biomarkers have their ALM within a small margin.
We estimate the ALM value of a vertex $x$ by using the ALM values of the rest of the vertices in $G$. 
Vertices in the neighborhood of $x$ in $G$ have a higher influence on its ALM. 
The influence of the vertices declines as we move farther from $x$ in $G$.
\subsection{Tug-of-War Equation}
Like supervised algorithms, $ p$-Laplacian-based regression needs a training set and a testing set.
We divide the vertex set, $\mathcal{X}$ into two subsets: $\Gamma$ and $\mathcal{X}\backslash \Gamma$. 
For vertices in $\Gamma$, we remember the values of the ALM for the corresponding patients, which makes $\Gamma$ our training set.
Similarly, for vertices in $\mathcal{X}\backslash \Gamma$, we temporarily forget the values of the corresponding ALM which makes the set $\mathcal{X}\backslash \Gamma$ our testing set.
Let $g : \Gamma\rightarrow \mathbb{R}$ be the true value of ALM for vertices in $\Gamma$ and $u:\mathcal{X}\rightarrow \mathbb{R}$ be our estimated value of ALM for vertices in $\mathcal{X}$. Clearly, $u$ and $g$ are identical in $\Gamma$.  

To estimate the value of ALM for the vertices in $\mathcal{X}\backslash \Gamma$, we solve the Game-Theoretic $p$-Laplacian for the function $u$ on the set $\mathcal{X}\backslash \Gamma$.
Hence, the following set of equations describes our system.
\begin{equation}
\label{eqn:p-Laplacian}
   \begin{cases}
    \mathcal{L}_pu(x)=0 &\text{if }x\in\mathcal{X}\backslash \Gamma\\
    u(x)=g(x) &\text{if } x\in \Gamma
\end{cases}
\end{equation}

In Equation \ref{eqn:p-Laplacian}, 
$p$ is in $[2,\infty)$.
To solve, $\mathcal{L}_p u(x)=0$, we utilize the Dynamic Programming Principle (DPP) described in \cite{Calder-Drenska}. DPP can be represented by the following equation:
\begin{equation}
\label{eqn:dpp}
    u(x)=\frac{\alpha}{d_x}\sum_{y\in N_x}w_{xy}u(y) +\frac{1-\alpha}{2}\bigg(\min_{N_x}u +\max_{N_x}u\bigg)
\end{equation}
where, $\alpha=\frac{1}{p-1}$ and $d_x=\sum_y w_{xy}$ is the degree of vertex $x$. 
\newline
Equation \ref{eqn:dpp} arises from a stochastic two-player tug-of-war game. The details of the procedure can be found in \cite{Calder-Drenska}.

In Equation \ref{eqn:dpp}, we observe when $p$ is 2, $\alpha $ is 1, which results in the elimination of the terms associated with the minimum and maximum values of $u$. Consequently, the model relies exclusively on the random walk. For higher values of \( p \), the contribution of the summation term diminishes.  Moreover,  as $p$ becomes larger, the factor $\frac{1-\alpha}{2}$ gets closer to $\frac{1}{2}$. 

We use a Python package \textit{GraphLearning} from \cite{Calder-github} to predict ALM. The algorithm first computes the weight matrix. The array \textit{X\textunderscore combined} stores the measurements of the  44 biomarkers for each patient.
For each pair of vertices $x_i$ and $x_j$ in \textit{X\textunderscore combined}, the matrix $W$ stores the weight $w_{x_iy_j}$.
The function \textit{gl.graph} then creates a $k$-NN graph from $W$.

\begin{lstlisting}[language=Python, basicstyle=\ttfamily\small, keywordstyle=\color{blue}, commentstyle=\color{gray}\itshape]
W = gl.weightmatrix.knn(X_combined, k)
G = gl.graph(W)

\end{lstlisting}

Now, the algorithm  runs the $p$-Lapalcian based regression.
Specifically,  the function \textit{ G.\text{plaplace}} computes \textit{yhat}, the predicted value of ALM based on the true values of ALM, \textit{y\textunderscore combined} of the training set. Clearly, \textit{yhat}, and \textit{y\textunderscore combined}, are equal for the vertices in the training set.
\begin{lstlisting}[language=Python, basicstyle=\ttfamily\small, keywordstyle=\color{blue}, commentstyle=\color{gray}\itshape]

yhat = G.plaplace(np.where(train_mask)[0], 
y_combined[train_mask], p)   
\end{lstlisting}





In Section \ref{subsec:results:p-Laplacian}, we discuss our results on $p$-Laplacian based regression.





\section{Results And Analysis} \label{sec:results}
In this section, we present our results for both supervised and semi-supervised algorithms. The RMSEs in this section are expressed as a percentage (see Equation \ref{eqn:RMSE}). The figures of heatmaps and line plots include nine subplots in a grid system. The columns and rows represent the datasets and the target variables respectively. 

\subsection{Polynomial Regression}
\label{subsec:results-polynomial}
Polynomial regression models (Traditional, Lasso, Ridge, and Bayesian Ridge) are compared across polynomial degrees (1, 2, and 3) and datasets. A key issue with traditional regression is its tendency to suffer from high variance, which increases as the polynomial degree increases. Regularization methods, such as Ridge and Bayesian Ridge, mitigate this variance to some extent, but overall, increasing the polynomial degree generally leads to higher errors.
\\

\begin{figure}[h!]
    \centering
    \includegraphics[width=1.0\linewidth]{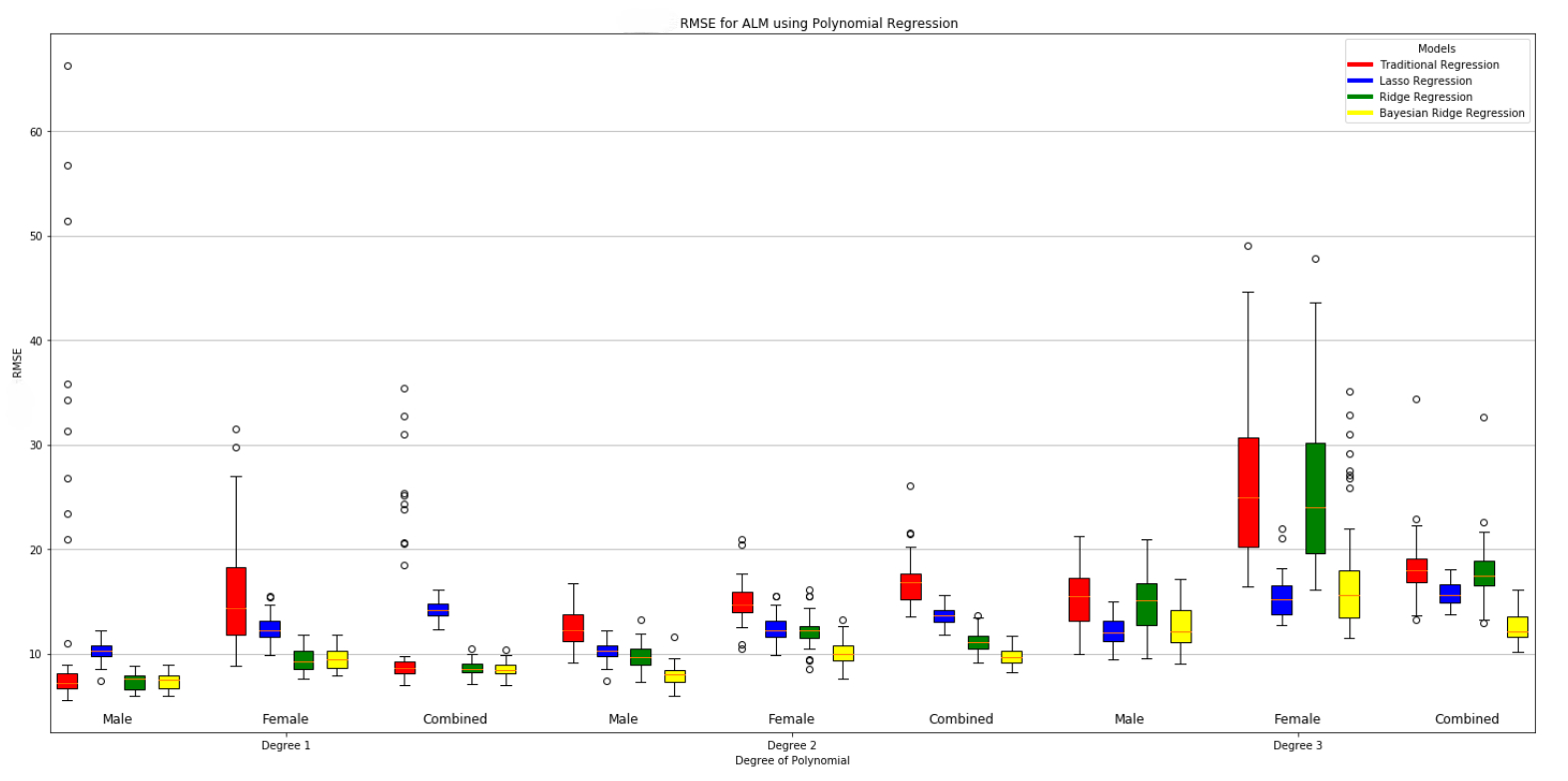}
    \caption{Distribution of RMSEs in predicting ALM by Polynomial Regression Models across Datasets }
    \label{appendix-fig-regression-alm}
\end{figure}

As Figure \ref{appendix-fig-regression-alm} exhibits, in the case of ALM Bayesian Ridge and Ridge Regression consistently achieve lower RMSEs (8–10 for degree 1, 10–15 for degree 2 and 15–20 for degree 3) across datasets. Traditional Regression shows higher variability, particularly for Female  with RMSEs exceeding 40–50 for higher-degree models.

\begin{figure}[h!]
    \centering
    \includegraphics[width=1.0\linewidth]{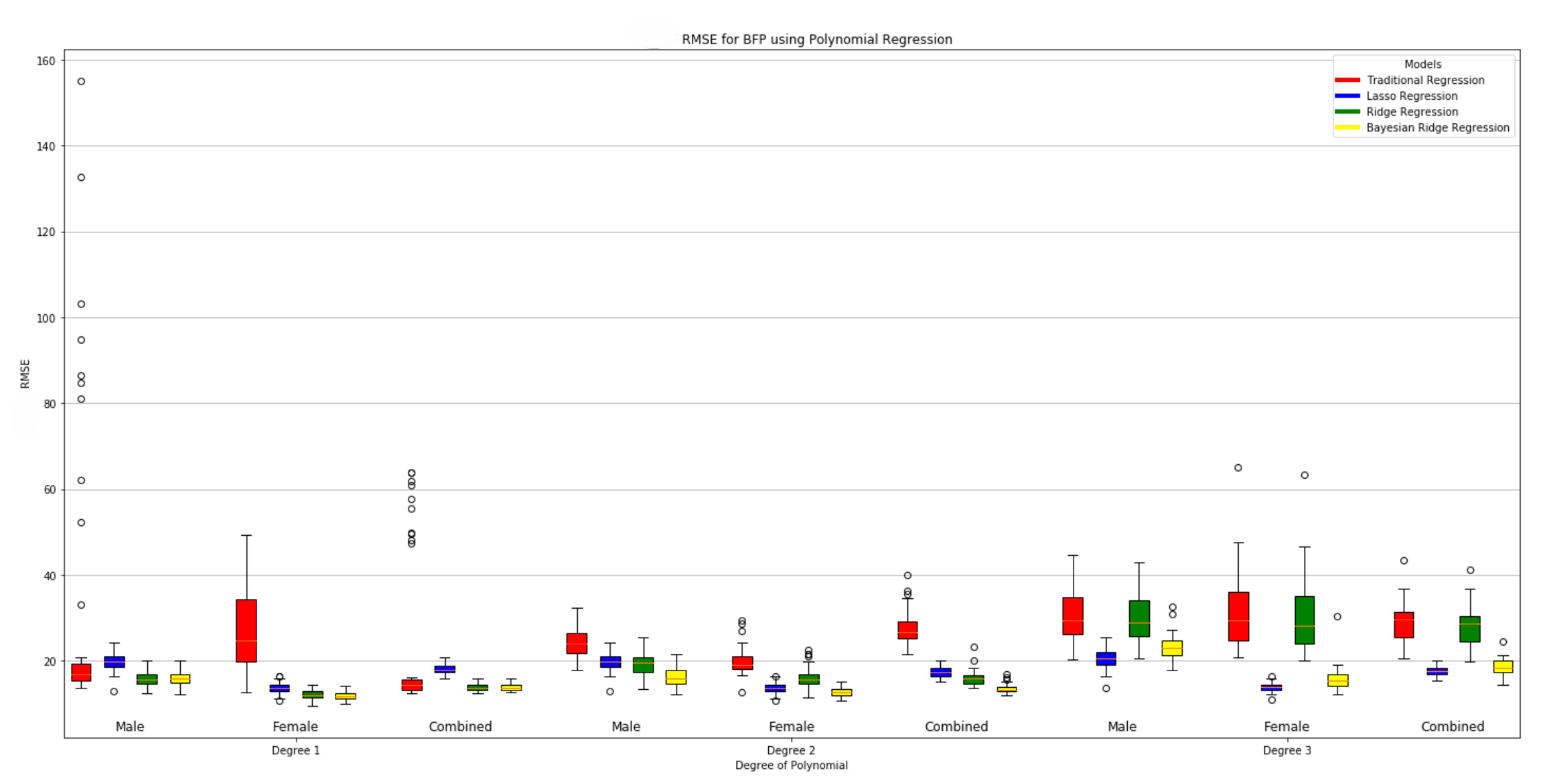}
  \caption{Distribution of RMSEs in predicting BFP by Polynomial Regression Models across Datasets  }
\label{appendix-fig-regression-bfp}
\end{figure}

Similarly, Figure \ref{appendix-fig-regression-bfp}d demonstrates Bayesian Ridge and Ridge Regression outperform the Traditional one for BFP, with RMSE between 8–20 across degrees. In contrast, Traditional Regression shows significant variability, especially for degree 3, where RMSEs exceed 40-60 for Female (Figure ~\eqref{appendix-fig-regression-bfp}).

Finally, Bayesian Ridge (See  Figure \ref{appendix:p-lap-2-BMD}) maintains the lowest RMSEs  for BMD (8–20 across degrees) compared to other methods. Traditional regression struggles with variability, particularly for Male and Female, with RMSEs reaching 60–70 for degree 3. 

\begin{figure}[ht!]
    \centering
    \includegraphics[width=1.0\linewidth]{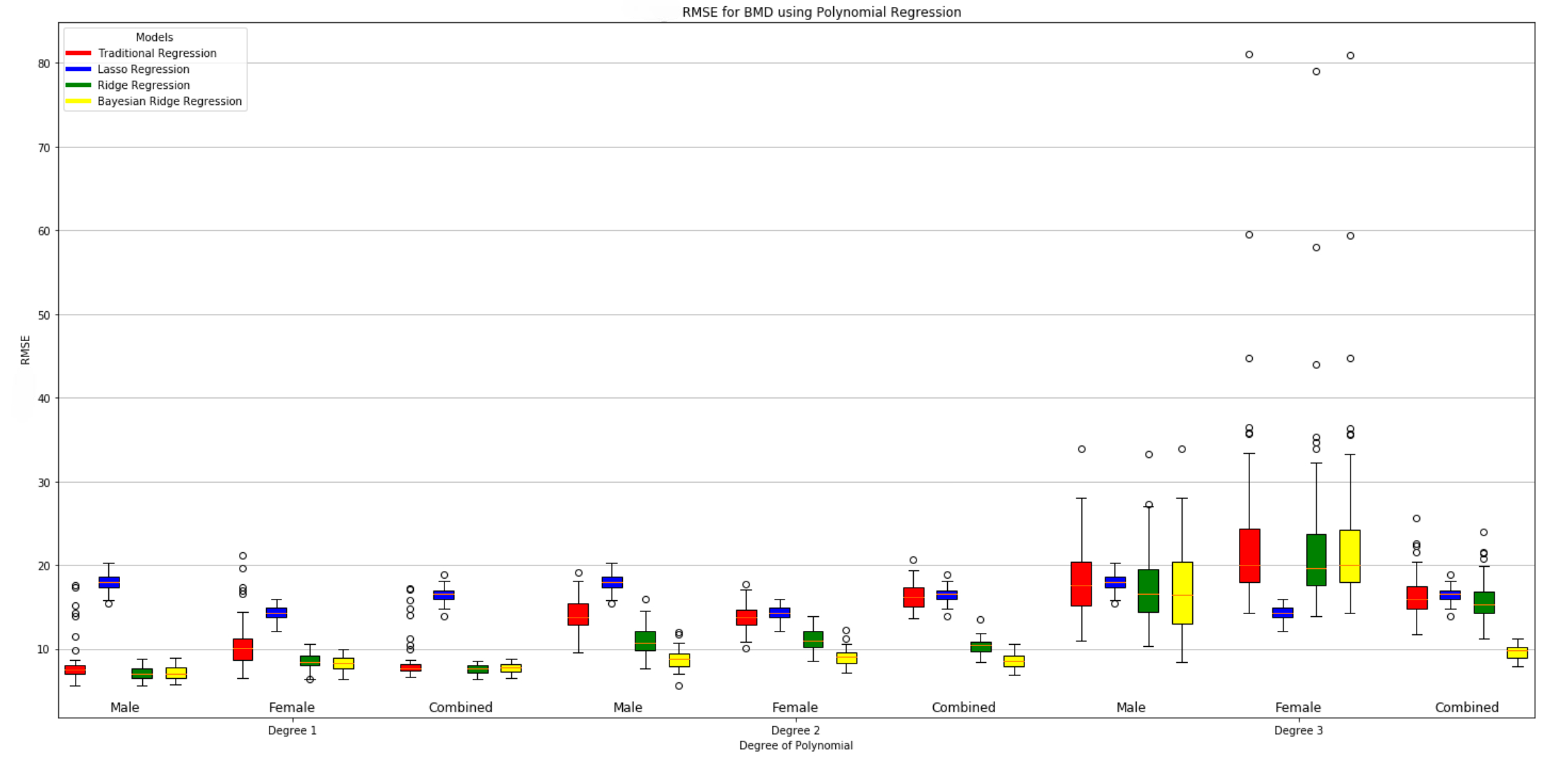}
      \caption{Distribution of RMSEs in predicting BMD by Polynomial Regression Models across Datasets}
    \label{appendix-fig-regression-bmd}
\end{figure}

Bayesian Ridge remains the most reliable method in all data sets and the degrees (Figure ~\eqref{appendix-fig-regression-bmd}).

\subsection{SVR and LSSVR}
\label{subsec:results:SVR-LSSVR}
For SVR with linear kernrel , we optimize the regularization parameter, $C$, and the margin tolerance, $\epsilon$, for predicting BMD, BFP, and ALM.\\
\begin{figure}[h]
    \centering
    \includegraphics[width=1.0\linewidth]{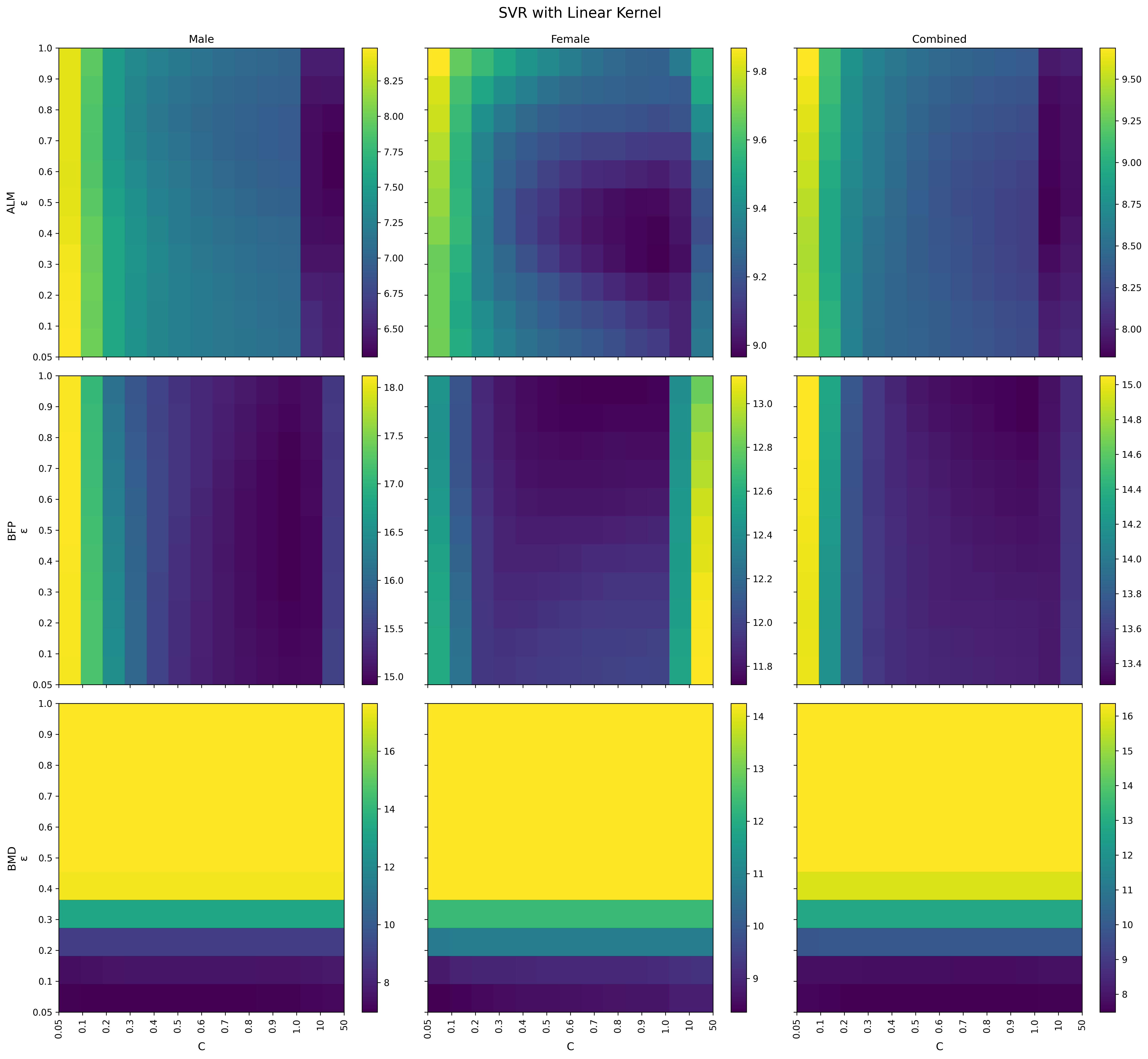}
    \caption{Heatmap of RMSEs (in \%) for SVR with parameters $C$ and $\epsilon$}
    \label{fig:heatmap:svr}
\end{figure}

As shown in Figure~\ref{fig:heatmap:svr}, SVR, at its best, yields error between  6\% and 9\% for ALM and BMD across all datasets, with RMSE values ranging between 6\% and 9\%.
The models performs better for Male and Combined while predicting these two biometrics. For these two biometrics, the model performs better for Male and Combined than for Female.
The model achieves optimal performance at higher values of \( C \) and \( \epsilon \) for Male ALM and Combined ALM compared to Female ALM.
When predicting BMD, the model optimizes at $\epsilon=0.05$ and $C$ in $[0,1]$ yielding between 6-9\% for all datasets.
Like the regression models, SVR performs poorly for BFP specially, for Male yielding errors close to 15\%.

\begin{figure}[h]
    \centering
    \includegraphics[width=1.0\linewidth]{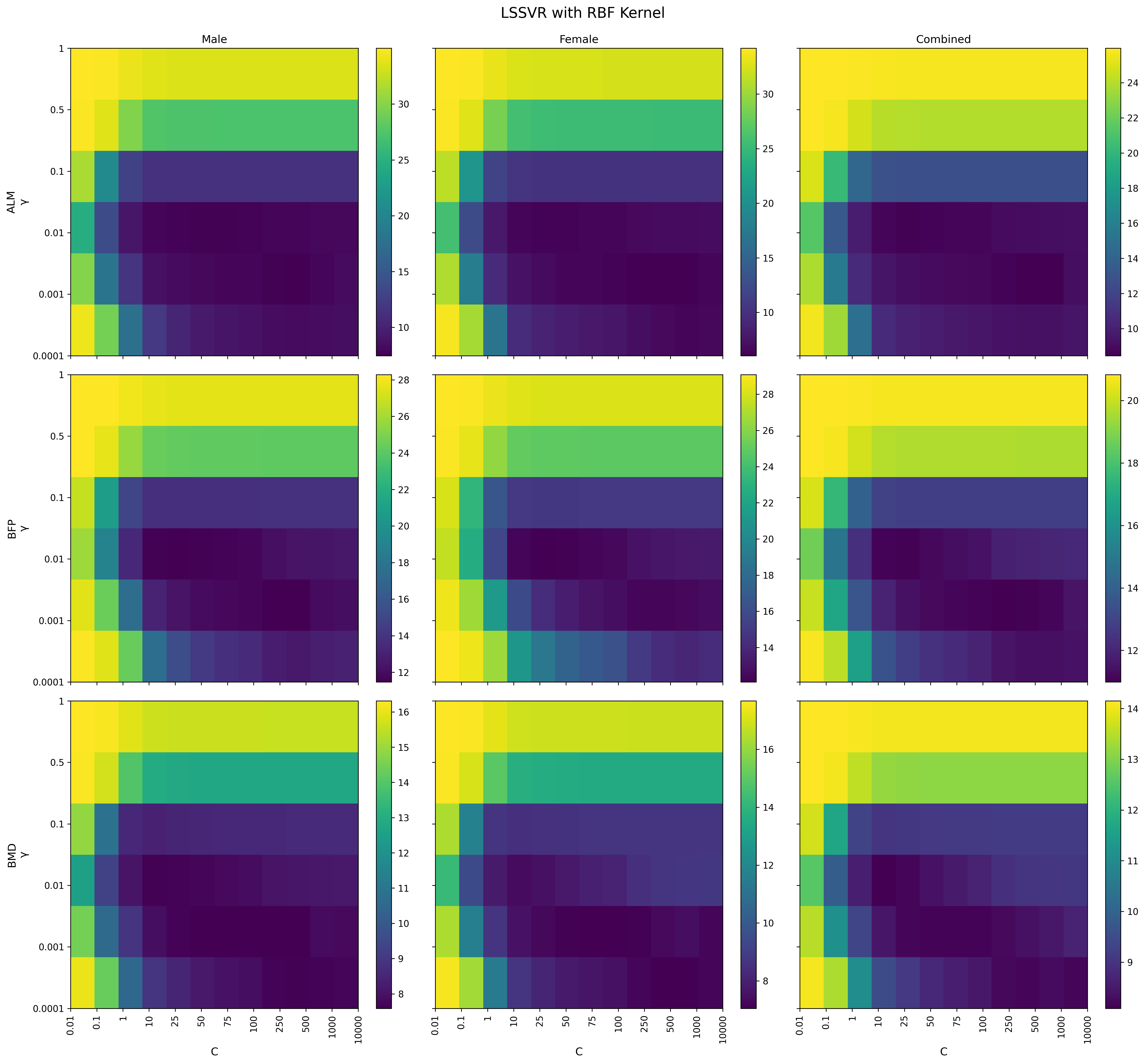}
    \caption{Heatmap of RMSEs (in \%) for LSSVR with parameters $C$ and $\gamma$}
    \label{fig:heatmap:lssvr}
\end{figure}

The heatmap in Figure~\ref{fig:heatmap:lssvr} demonstrates the performance of LS-SVR with an RBF kernel across ALM, BFP, and BMD datasets for varying values of $C$ (regularization) and $\gamma$ (kernel parameter).
 
For ALM, the best performance was observed with $C$ in the range of 500 to 1000 and $\gamma=0.001$ achieving RMSE values between 6 \% and 8 \% across all datasets. For BFP, the optimal results were obtained with $C$ in the range of 25 to 250 and $\gamma$ between 0.001 and 0.01, yielding RMSE values between 11 \% and 13 \%. For BMD, the best-performing region had $C$ values from 10 to 250 and $\gamma$ between 0.001 and 0.01, with RMSE values ranging from 7\% to 8\%. These trends suggest that lower $\gamma$ values combined with higher $C$ values provide better accuracy for ALM and BMD, while BFP benefits from moderate $\gamma$ and smaller $C$ values.






\subsection{Random Forest and XGBoost}
\label{subsec:Results:RF-XGBoost}
For RF and XGBoost models, we varied the number of decision trees, $n$, and their maximum depth, $d$.\\
\begin{figure}[h]
    \centering
    \includegraphics[width=1.0\linewidth]{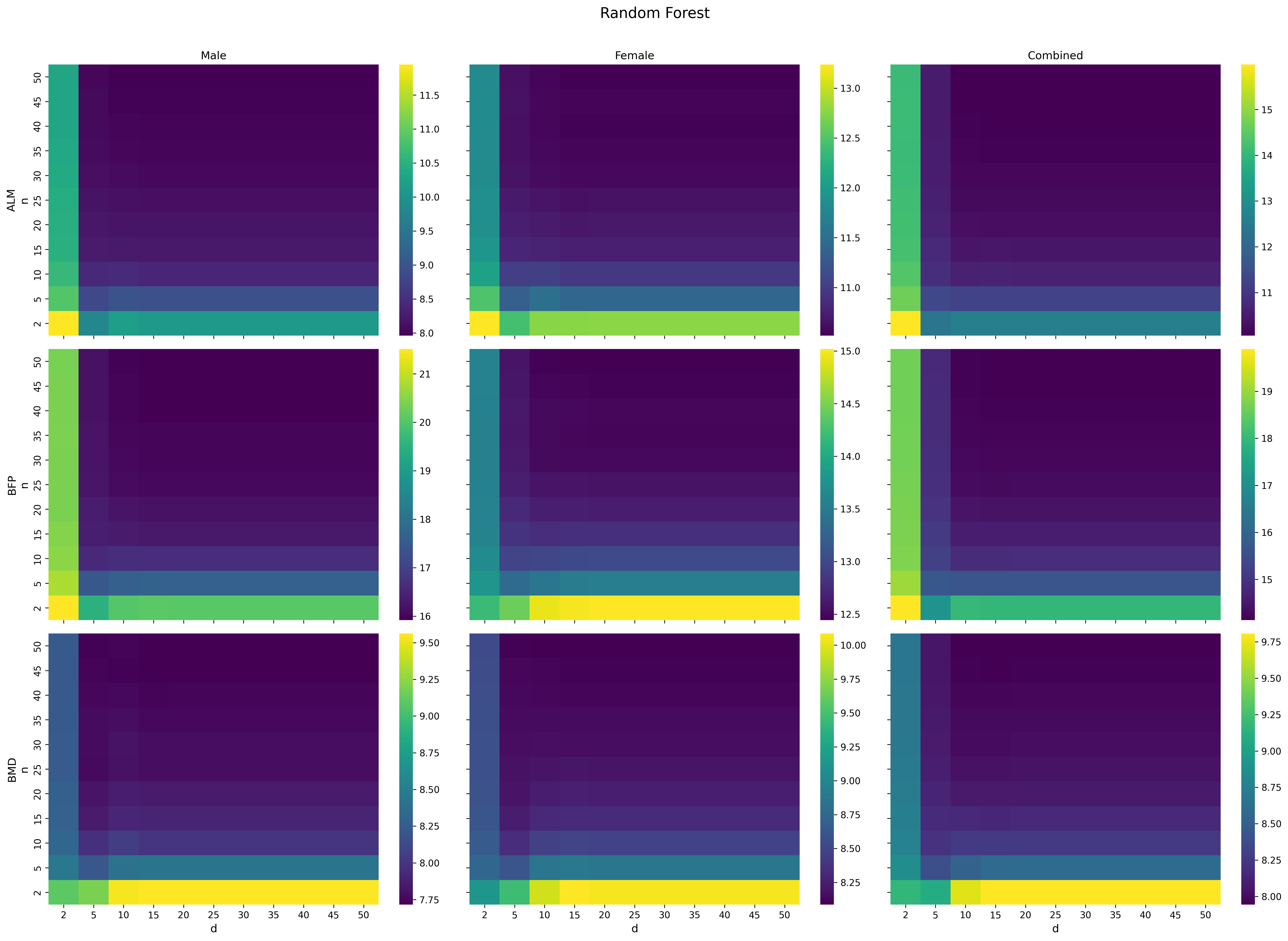}
    \caption{Heatmap of RMSEs (in \%) for RF with parameters $n$ and $d$ }
    \label{fig:heatmap:rf}

\end{figure}
\begin{figure}[h]
    \centering
    \includegraphics[width=1.0\linewidth]{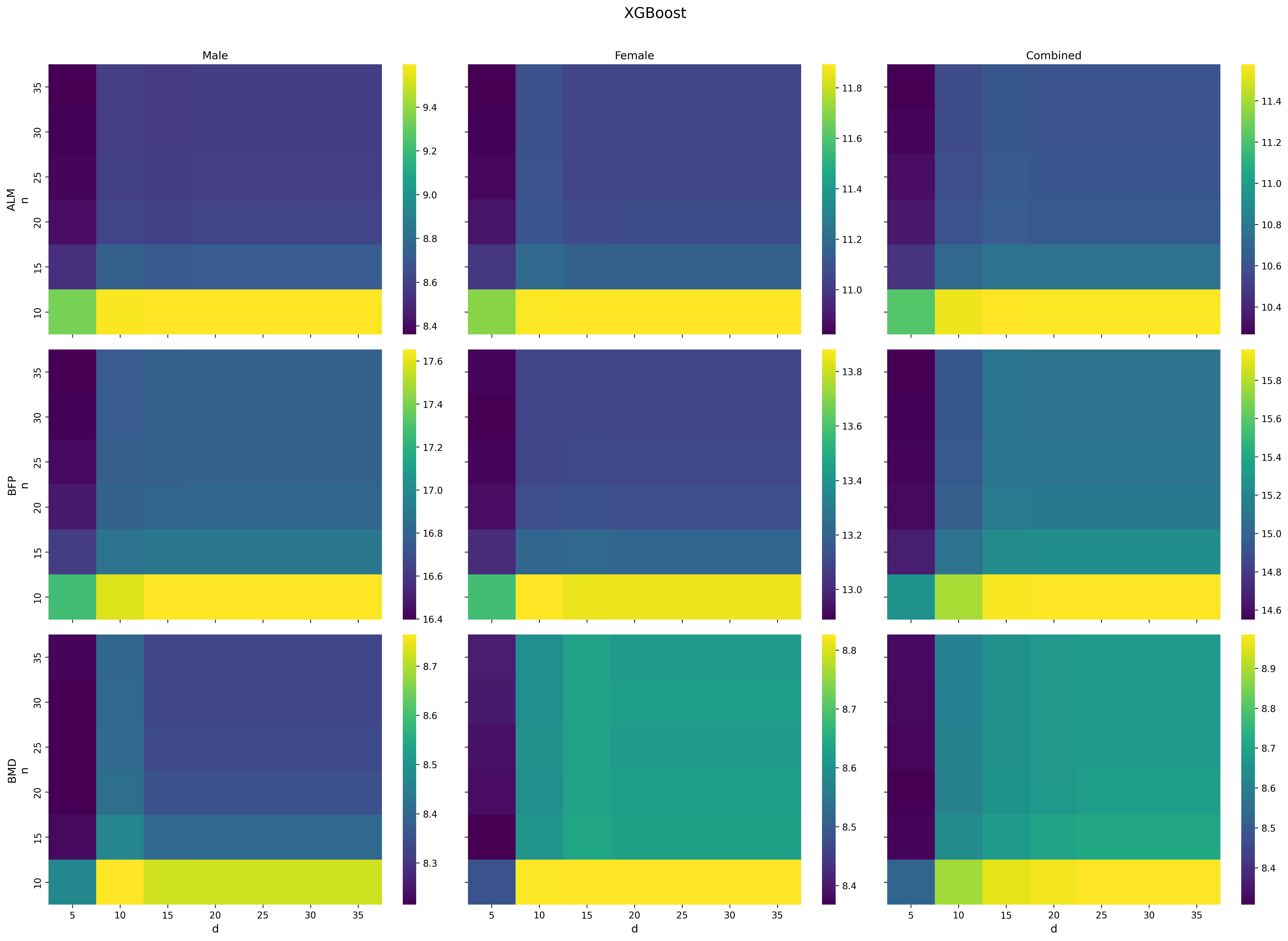}
    \caption{Heatmap of RMSEs (in \%) for XGBoost with parameters $n$ and $d$ }
    \label{fig:heatmap:xgboost}
\end{figure}
For RF, $n$ and $d$ were between 2 and 50.
As we see in Figure~\ref{fig:heatmap:rf}, RF performs the best for BMD and the worst for BFP. For BMD, the RMSE for Male, Female, and Combined are 7.72\% ($n=45,d=15$), 8.09\% ($n=50,d=20$) and 7.94\% ($n=50,d=10$), respectively. 
For ALM, RF achieves the lowest RMSEs within the range of 7 to 11 percent for all the datasets.
The model performs the worst for Male BFP as the lowest RMSE for all $n$ and $d$ it yields is 15.92\%.
As demonstrated by Figure \ref{fig:heatmap:rf}, the RMSE stays relatively constant for all datasets and target variables when $n\geq 25$ and $d\geq 15$,

Similar to  RF, the RMSE of XGBoost exhibits minimal variation once $n$ and $d$ exceed a certain threshold. We tested the model for the values of $n$ and $d$  between 5 and 35. The model's run time is exceptionally high for larger values of the parameters.

The model performs the best for BMD by yielding errors between 8 and 9 percent for each dataset. 
The RMSE for BMD for Male, Female and Combined are 8.22 \% ($n=20,d=6)$, 8.37\%($n=15,d=5)$, and 8.31\%($n=20,d=5$) respectively. 
Like RF and the other models, its performance is the worst for BFP. In all three datasets and target variables, the model yields the highest lower limit error of 16.4 \% for Male BFP.
The model's performance is similar to RF for ALM with RMSE ranging between 8-12\%.
As Figure~\ref{fig:heatmap:xgboost} demonstrates, the most optimal $d$ for XGBoost is $5$ for all datasets and target variables. RMSE only increases when $d>5$.

\subsection{Neural Network}
\label{subsec:results:NN}
\begin{figure}[h!]
    \centering
    \includegraphics[width=1.0\linewidth]{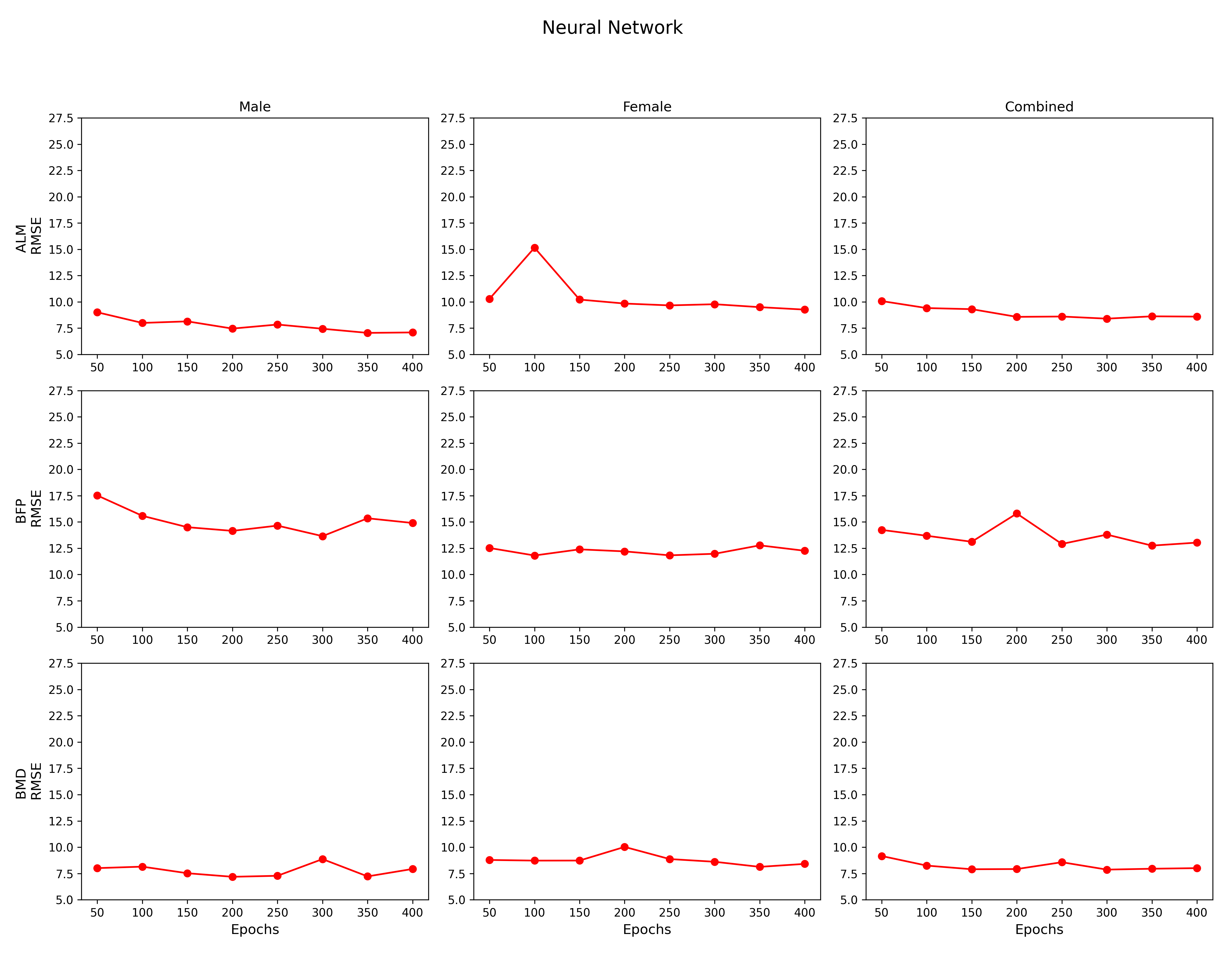}
    \caption{RMSE of Neural Network vs Number of Epochs}
    \label{fig:lineplot-}
\end{figure}
As explained in Section \ref{subsec:supervised-nn}, we experimented with multiple activation functions, batch sizes, and optimizers. 
Here, we show how our neural network performed for different number of epochs. 
For Combined,
 as training epochs increase, RMSE for ALM, BFP, and BMD predictions drops.
At $50$ epochs, RMSE is higher for all three, but it stabilizes around $8.8$ for ALM, $13.5$ for BFP, and $8.0$ for BMD at $400$ epochs.
This shows the model's accuracy improves with more training, leading to better predictions.

For Female, ALM's RMSE drops from 10.5 to 9.0 as epochs increase from 50 to 400. BFP's RMSE improves significantly, falling from 15.0 to 13.0. BMD's RMSE reduces from 9.0 to 8.0. More training epochs boost accuracy for all metrics in female data.

In  Male, ALM's RMSE decreases from 9.5 to 7.5 with more epochs. BFP starts high at 17.0 but drops to 14.0. BMD's RMSE falls from 9.0 to 7.5 as the number of epochs increases. More epochs lead to better performance, especially for ALM and BFP predictions.

Overall, RMSEs consistently decrease with more epochs, showing improved predictive accuracy. This trend is more pronounced in male data. 

\subsection{p-Lapacian based Regression}
\label{subsec:results:p-Laplacian}
We tested the $p$-Laplacian based regression under two schemes and training-testing ratios.
As described in Section \ref{sec:semi-supervised-alg},the model is implemented on a graph where the nodes are the patients and the edges are constructed based on the similarity between the patients.
We constructed the graph using two methods. In the first method, we used all 44 biomarkers to define similarity and then construct the graph. In the second method, we used the top ten biomarkers listed in Tables \ref{appendix:top-10-bio-ALM},  \ref{appendix:top-10-bio-BFP}, and  \ref{appendix:top-10-bio-BMD} for each target variable and dataset. The idea behind this is to see if using fewer and more significant biomarkers only improves the model's performance.
We call the first method $p$-Laplacian-1 and the second method $p$-Laplacian-2. 
Like the supervised models, $p$-Laplacian-1 and $p$-Laplacian-2 are first implemented with $K$-cross validation for $K=5$.
To test the model's performance when trained on a smaller percentage of training data, we modified the 
$K$-fold cross-validation slightly. We used one fold as our training set and the other four as our testing set.
We tested both versions of the model using the modified $K$-fold cross-validation for $K=2,3,4,5,10,$ and $20$ with training testing ratios 50-50, 33-67,25-75, 20-80, 10-90, and 5-95, respectively.

In the following discussion, we represent the percentage of training data as Training \%.
We vary two parameters: $p$ (see the description of Equation \ref{eqn:dpp}) and $k$ (the number of neighbors every node has in the graph).
For optimization, we implemented
$p$-Laplacian-1 and $p$-Laplacian-2 for different combinations of $p$ and $k$ for a fixed Training \%. 
For both models, the optimal $p$ is at most 10 and the optimal $k$ is at most 60.
Figure \ref{fig:p-Laplacian-1,2-RMSE} demonstrates how the lowest RMSE $p$-Laplacian-1 and $p$-Laplacian-2 yielded changes with Training \%. 
The RMSEs of both models increase as the Training \% decreases. This phenomenon is not unusual. Calder and Drenska \cite{Calder-Drenska}, showed for a random geometric graph like $k$-NN graph the accuracy of the $p$-Laplacian-based algorithm can decline with label rate which is equivalent to Training \%.
There is a relatively steep increase in the models' RMSE  when the Training \% is reduced from 10 to 5 across all target variables and datasets. 
Among the target variables, they both
perform the best for BMD. The yielded RMSE is less than 10.2\% for all datasets when the Training \% is less than or equal to
10.
The RMSEs of both models follow similar trajectories for BMD.
For Male ALM and Female ALM, $p$-Laplacian-2 does slightly better than $p$-Laplacian-1; however, for Combined ALM, it yields higher RMSEs than $p$-Laplacian-1 when Training \% is at least $10$.
Like the supervised models, $p$-Laplacian-1 and $p$-Laplacian-2 perform poorly on estimating BFP, especially for Male and Combined. $p$-Laplacian-1 outperforms $p$-Laplacian-2 by a huge margin when estimating BFP for these two datasets.
Both $p$-Laplacian-1 and $p$-Laplacian-2 perform well for BMD yielding errors below $10$\% for all the datasets even when they are trained on only $10$ percent of the the data. The errors are slightly above $10$\% when Training \% is reduced to $5$\%.

Figure \ref{fig:p-Laplacian-1,2-p} and Figure \ref{fig:p-Laplacian-1,2-k} demonstrate how the optimal parameters $p$ and $k$ change for $p$-Laplacian-1 and $p$-Laplacian-2, respectively with the Training \%. 
As illustrated by Figure \ref{fig:p-Laplacian-1,2-p}, 
when Training \% is 10 or 5, the optimal $p$ values for $p$-Laplacian-1 and $p$-Laplacian-2 jump above 2 for ALM and BMD. In the case of BFP, this happens for $p$-Laplacian-1 but not for $p$-Laplacian-2.
Calder \textit{et al.} \cite{calder2023rates} proved at low label rates, (equivalent to  Training\%)  a  2-Laplacian algorithm ( when $p=2$) becomes degenerate which implies it fails to predict the target variable with high accuracy.
Another observation is that $p$-Laplacian-1 is optimized at much higher $p$ values when compared to $p$-Laplacian-2. 
We did not find any evidence on why $p$-Laplacian-2 acts as an anomaly for BFP.
One reason could be that weakly correlated biomarkers with the target variables make the convergence of the RMSEs slower. Further investigation is required to substantiate this hypothesis.

Figure \ref{fig:p-Laplacian-1,2-k} demonstrates the optimal $k$ values for both $p$-Laplacian models. There is no obvious pattern to distinguish between the optimal $k$ values of $p$-Laplacian 1 and $p$-Laplacian-2.

\begin{figure}[h!]
    \centering
    \includegraphics[width=1.0\linewidth]{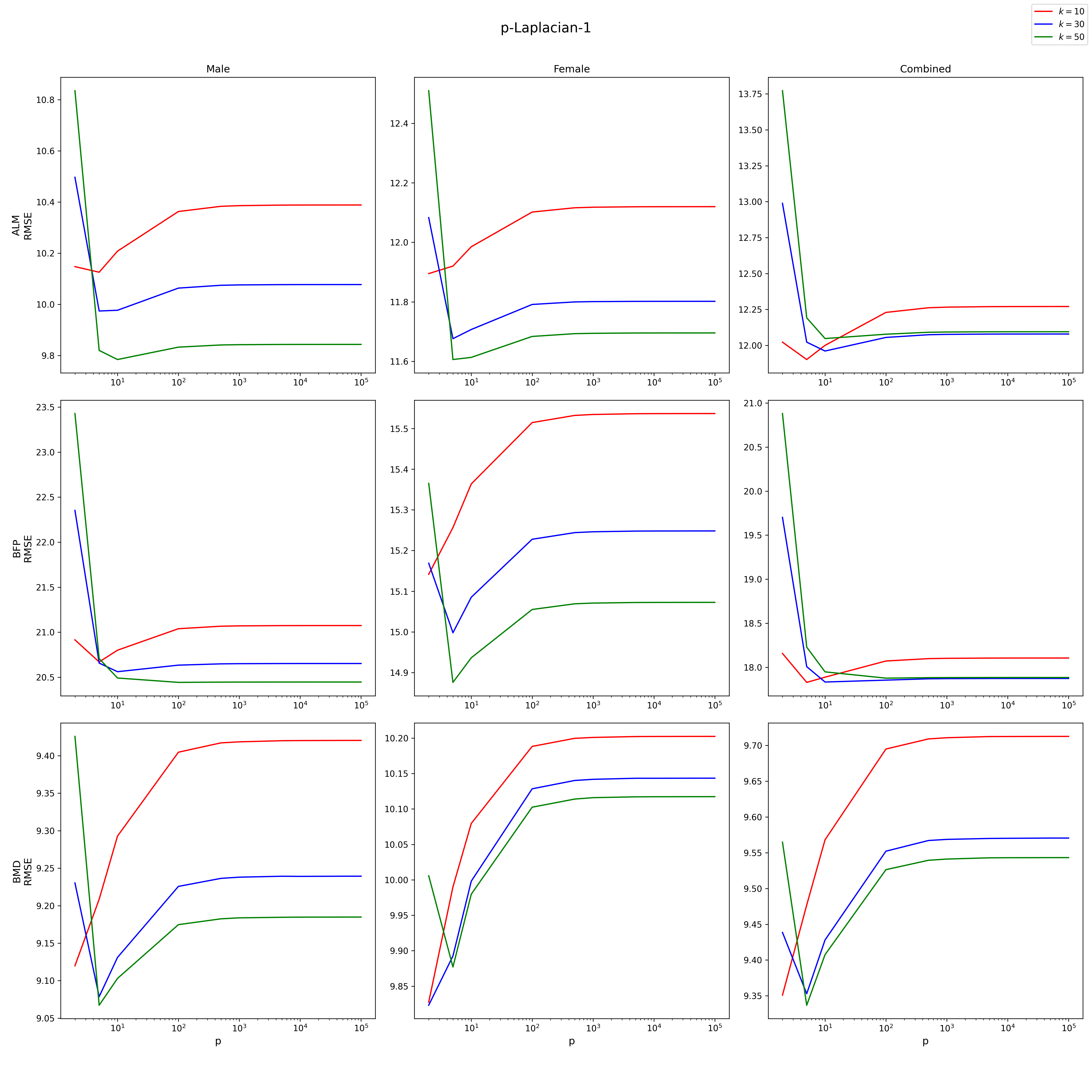}
    \caption{Asymptotic behavior of $p$-Laplacian-1 as $p\rightarrow \infty$ for $k=10,30$ ,and $50$}
    \label{fig:p-laplacian-1-asymptotic}
\end{figure}

\begin{figure}[h!]
    \centering
    \includegraphics[width=1.0\linewidth]{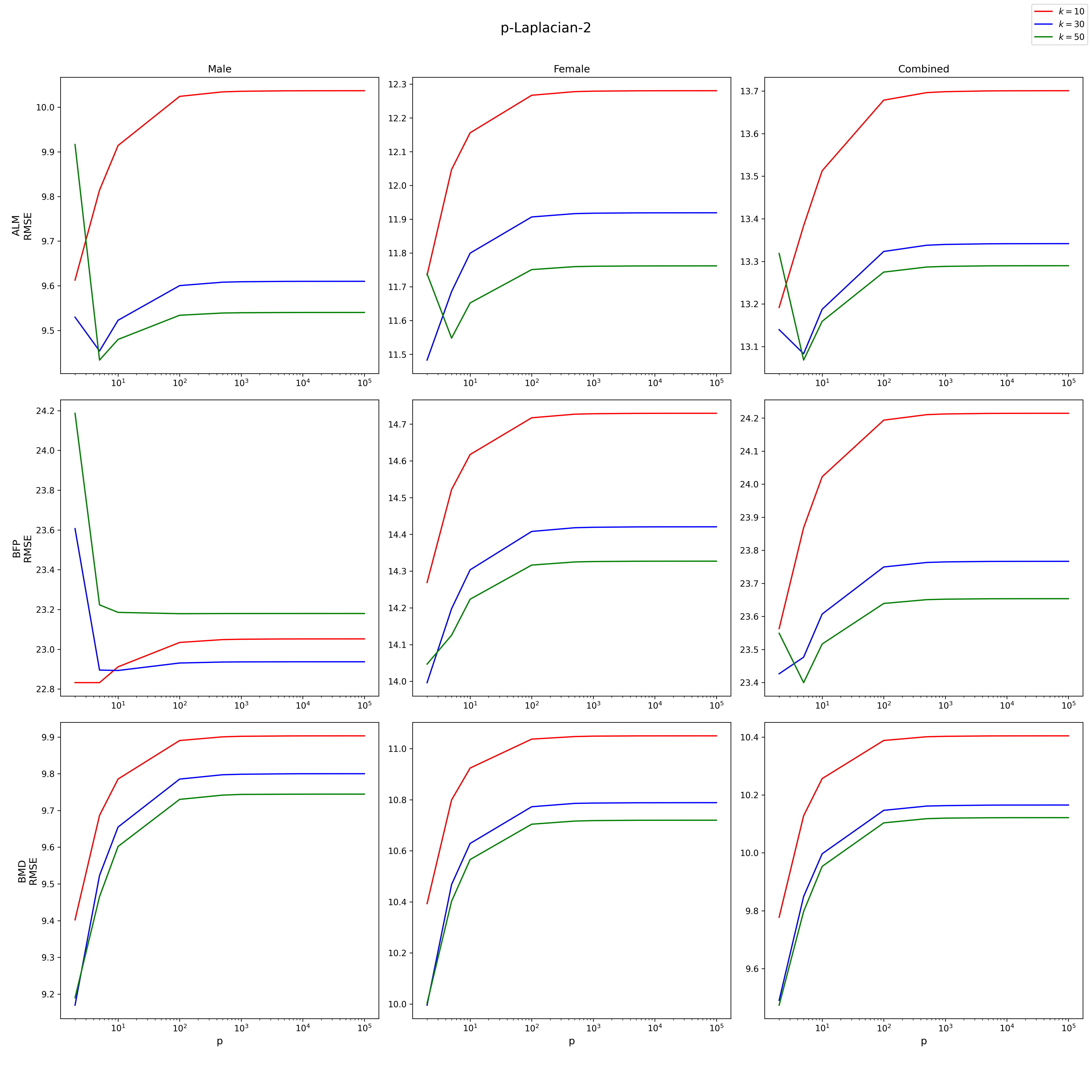}
    \caption{Asymptotic behavior of $p$-Laplacian-2 as $p\rightarrow \infty$ for $k=10,30$, and $50$}
    \label{fig:p-laplacian-2-asymptotic}
\end{figure}

Since $p$ can approach infinity, it is natural to investigate the asymptotic behavior of $p$-Laplacian based regression.
Equation \ref{eqn:dpp} predicts that the RMSE of a $p$-Laplacian based regression should converge for larger values of $p$. 
To verify this prediction, we implemented $p$-Laplacian-1 and $p$-Laplacian-2 for larger values of $p$ by fixing $k=10, 30$, and $50$.
Figure \ref{fig:p-laplacian-1-asymptotic} and Figure \ref{fig:p-laplacian-2-asymptotic} demonstrate the asymptotic behavior of RMSEs of $p$-Laplacian-1 and $p$-Laplacian-2 respectively when $p$ approaches infinity. 
For demonstration, Training \% is fixed at 20\%; however, similar results are achieved for other values of Training \%.
As Figure  \ref{fig:p-laplacian-1-asymptotic} and Figure \ref{fig:p-laplacian-2-asymptotic}  demonstrate, the RMSEs converge as $p$ gets larger.


\begin{figure}[h!]
    \centering
    \includegraphics[width=1.0\linewidth]{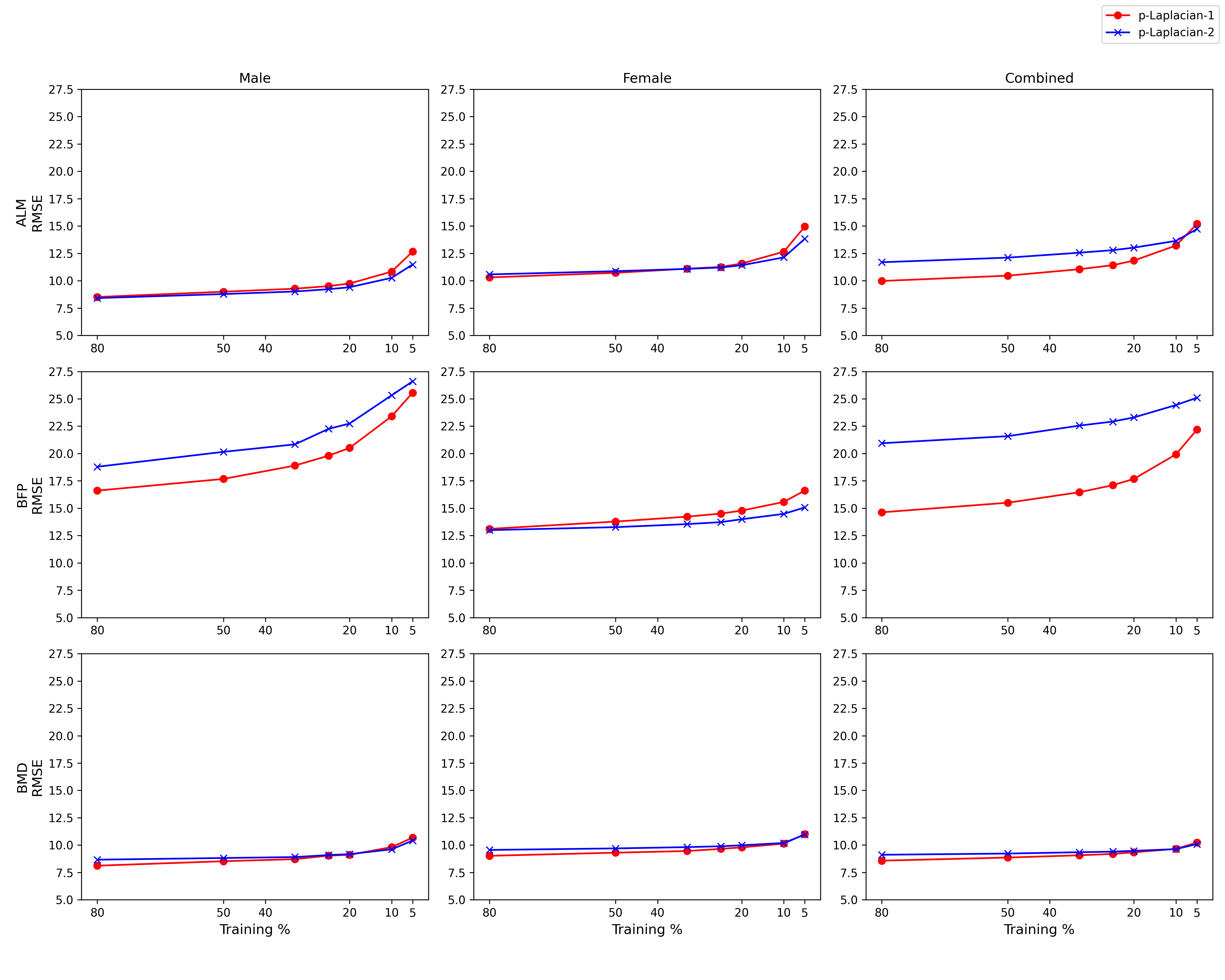}
    \caption{RMSEs (in \%) vs Training \% of $p$-Laplacian-1 and $p$-Laplacian-2 across the target variables and datasets}
    \label{fig:p-Laplacian-1,2-RMSE}
\end{figure}
\begin{figure}[h!]
    \centering
    \includegraphics[width=1.0\linewidth]{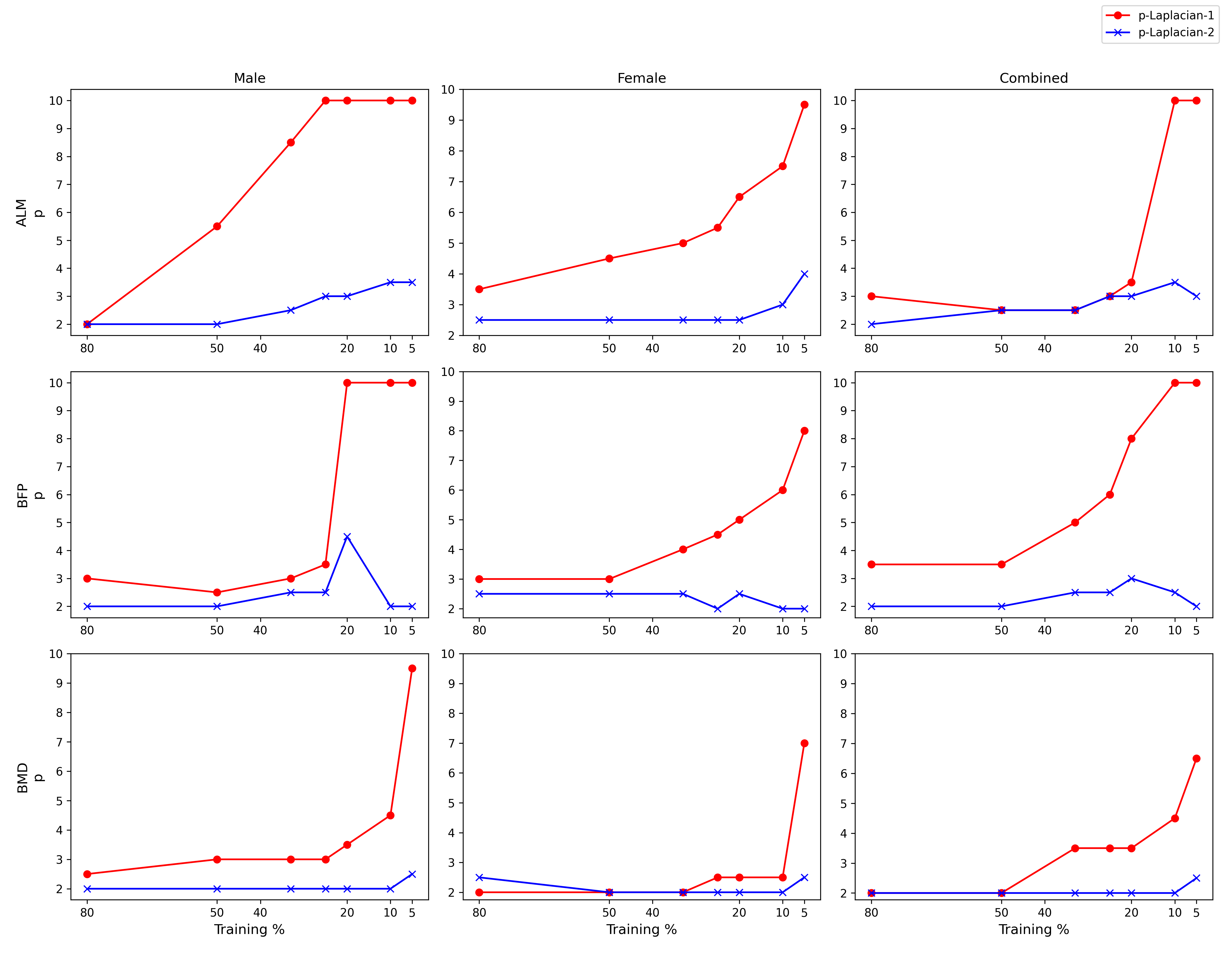}
    \caption{Optimal $p$ vs Training \% of $p$-Laplacian-1 and $p$-Laplacian-2 across target variables and datasets}
    \label{fig:p-Laplacian-1,2-p}
\end{figure}
\begin{figure}[h!]
    \centering
    \includegraphics[width=1.0\linewidth]{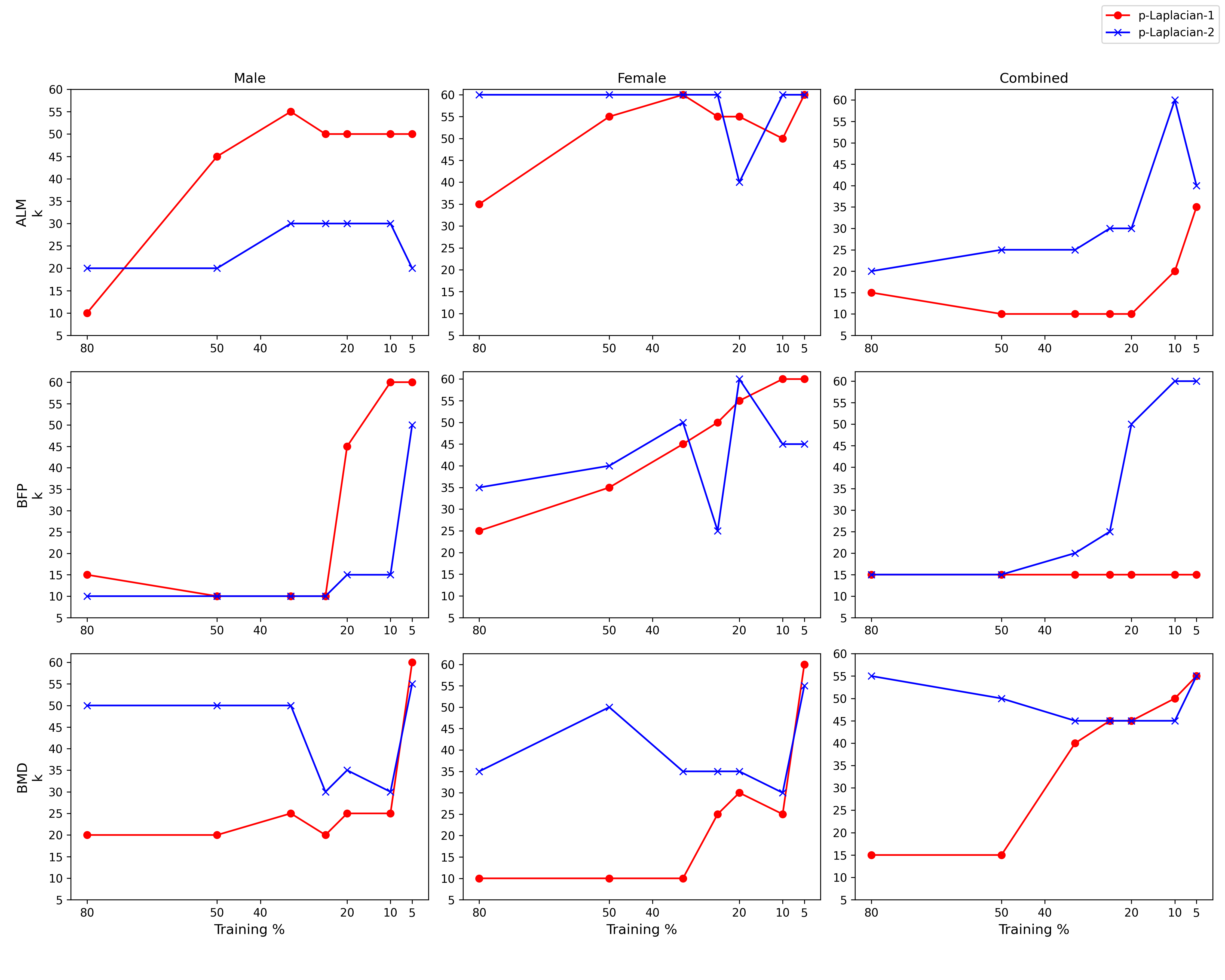}
      \caption{Optimal $k$ vs Training \% of $p$-Laplacian-1 and $p$-Laplacian-2 across target variables and datasets}
    \label{fig:p-Laplacian-1,2-k}
\end{figure}

.




\section{Discussion}
\label{sec:discussion}
There are two main objectives this article. The first is to test if ALM, BFP, and BMD can be predicted by applying supervised and semi-supervised algorithms on biomarkers' measurements obtained by the procedure described in Section \ref{sec:Data-Extraction}.
We used six different supervised algorithms namely-Linear/Polynomial Regression, SVR, LSSVR, Random Forest, XGBoost and Neural Network.
The second objective is to see if we can reduce the data we need to train a model to predict these three quantities. 
To achieve this, we used $p$-Laplacian based regression, a semi-supervised algorithm.

The supervised models demonstrated significantly higher accuracy for ALM and BMD compared to BFP.  The BFP predictions for Female were on average better than those for Male and Combined . One of the reasons behind models' poor performance on predicting BFP could be its relatively low correlations with the biomarkers. 
The  SVR and LSSVR models are the most accurate models for all target variables across all datasets. Here is the summary of the best performing supervised models for the target variables across all datasets with their respective RMSEs.


\begin{table}[ht]
\centering
\resizebox{\columnwidth}{!}{%
\begin{tabular}{l|cc cc cc}
\toprule
 & \multicolumn{2}{c}{\textbf{ALM}} & \multicolumn{2}{c}{\textbf{BFP}} & \multicolumn{2}{c}{\textbf{BMD}} \\
\midrule
\textbf{Male}     & SVR   & 6.30  & LSSVR & 11.47 & SVR   & 6.97  \\
\textbf{Female}   & LSSVR & 6.03  & SVR   & 11.72 & LSSVR & 7.05  \\
\textbf{Combined} & SVR   & 7.83  & LSSVR & 10.99 & SVR   & 7.48  \\
\bottomrule
\end{tabular}
}
\caption{Summary of the best performing Supervised Models  for  the Target variables  with respective RMSEs (in \%).}
\label{table:best-superivised-models}
\end{table}

Our results show that SVR and LSSVR have the potential to be an alternative to the DXA machine when measuring ALM and BMD.

Compared to the supervised regression algorithms we implemented in this paper, $ p$-Laplacian based regression is not so well studied.
In our analysis, $p$-Laplacian demonstrates that in the future, researchers could leverage using it, especially when training data is computationally or financially expensive. 
Like the supervised models, both $p$-Laplacian-1 and $p$-Laplacian-2 perform considerably better for ALM and BMD compared to BFP. 
The errors are below $10$\% even when the training percent is significantly low.
To predict BFP more accurately using $p$-Laplacian based regression, a higher correlated set of biomarkers with BFP could be used to construct the graph.

It is compelling to examine how the performance of the $p$-Laplacian model changes with the size of the dataset for a fixed proportion of training data. This way we can test if the number of data points in training impacts the model's performance differently than the proportion of training data.


Moreover, in our analysis, all the biomarkers are weighed equally to construct the similarity-based graph. Hence, one potential direction for future research is weighing the biomarkers differently. One such weighing could be based on how correlated they are with the target variable. 

In summary, our results show DEXA could be potentially replaced in the future to estimate ALM and BMD. However, further investigation is needed on larger datasets. Additionally, our investigation of the $p$-Laplacian model yielded positive results for ALM and BMD in reducing the amount of training data while maintaining a high level of accuracy. Our analysis opens new avenues to implement and analyze the $p$-Laplacian model in regression problems.

\section{Acknowledgements}

We would like to thank the following people for their contribution to the early mathematical stages of the project: Gowri Priya Sunkara, Iswarya Sitiraju, Alex P Mensen-Johnson, Alyssa M Blount, Amaya Evans, Andrew Regan, Anna S Crifasi, Artem Mukhamedzianov,  Atif Iqbal, Berend P Grandt, David Aiken, Evan Short, Gaurang Chauhan, Giovanni Ohashiegbula, Jamar K Whitfield, Kathyryn C Simon,  Nuwanthi Samarawickrama, Serene Sam, Seth Stephens, Sunella A Ramnath, Teddie E Swize, Tristan Desoto, Joseph Dorta. We would like to thank  Cassidy McCarty for her involvement with collecting the data. 

This work was partially supported by National Science Foundation DMS grant 2407839 as well as by National Institutes of Health NORC Center Grants P30DK072476, Pennington/Louisiana, P30DK040561, Harvard; and R01DK109008, Shape UP! Adults. This work was also supported by the Bodin Undergraduate Summer Research Fund in the LSU Department of Mathematics.

\bibliographystyle{plainnat}
\bibliography{references}

\vskip0.1in

%
\onecolumn

\newpage
\onecolumn


\newpage
\begin{table}[h!]
    \centering
    \resizebox{\textwidth}{!}{%
    \renewcommand{\arraystretch}{1.0}
    \begin{tabular}{lll}
\toprule

Height (cm) & Weight (kg) & Abdomen Circumference \\
Ankle Circumference Left & Arm Length Left & Arm Volume Left \\
Bicep Circumference Left & Calf Circumference Left & Chest \\
Collar Circumference & Forearm Circumference Left & Head Circumference \\
Hip Circumference & Horizontal Waist & Inseam Left \\
Leg Volume Left & MidThigh Circumference Left & Narrow Waist \\
Outside Leg Length Left & Seat Circumference & Surface Area Arm Left \\
Surface Area Leg Left & Surface Area Torso & Surface Area Total \\
Thigh Circumference Left & Torso Volume & Upper Arm Circumference Left \\
Volume & Waist Circumference & Ankle Circumference Right \\
Arm Length Right & Arm Volume Right & Bicep Circumference Right \\
Calf Circumference Right & Forearm Circumference Right & Inseam Right \\
Leg Volume Right & MidThigh Circumference Right & Outside Leg Length Right \\
 Surface Area Arm Right & Surface Area Leg Right & Thigh Circumference Right  \\
Upper Arm Circumference Right &Age  &  \\
\bottomrule
\end{tabular}
}
    \caption{List of the Biomarkers}
    \label{appendix-list-of-bio}
\end{table}

\begin{table}[ht]
\centering

\resizebox{\textwidth}{!}{%
\renewcommand{\arraystretch}{1.0}
\begin{tabular}{lclclc}
\toprule
\textbf{Male} & \textbf{Corr. Coeff.} & \textbf{Female} & \textbf{Corr. Coeff.} & \textbf{Combined} & \textbf{Corr. Coeff.} \\
\midrule
Surface Area Total & 0.964 & Surface Area Total & 0.907 & Surface Area Total & 0.910 \\
Surface Area Leg Right & 0.956 & Leg Volume Left & 0.899 & Surface Area Arm Left & 0.904 \\
Surface Area Leg Left & 0.953 & Surface Area Leg Right & 0.898 & Surface Area Arm Right & 0.902 \\
Surface Area Arm Left & 0.947 & Surface Area Leg Left & 0.897 & Weight (kg) & 0.888 \\
Leg Volume Left & 0.944 & Leg Volume Right & 0.894 & Surface Area Torso & 0.875 \\
Leg Volume Right & 0.943 & Weight (kg) & 0.891 & Forearm Circumference Right & 0.872 \\
Surface Area Arm Right & 0.939 & Volume & 0.868 & Surface Area Leg Right & 0.869 \\
Weight (kg) & 0.934 & Calf Circumference Left & 0.865 & Volume & 0.865 \\
Volume & 0.922 & Thigh Circumference Left & 0.862 & Surface Area Leg Left & 0.865 \\
Surface Area Torso & 0.916 & Thigh Circumference Right & 0.861 & Arm Volume Left & 0.861 \\
\bottomrule
\end{tabular}%
}
\caption{Top 10 most correlated biomarkers with ALM with their Corresponding Correlation Coefficient}
\label{appendix:top-10-bio-ALM}
\end{table}

\begin{table}[h!]

\centering
\resizebox{\textwidth}{!}{%
\begin{tabular}{lclclc}
\toprule
\textbf{Male} & \textbf{Corr. Coeff.} & \textbf{Female} & \textbf{Corr. Coeff.} & \textbf{Combined} & \textbf{Corr. Coeff.} \\
\midrule
Horizontal Waist & 0.577 & Horizontal Waist & 0.754 & Waist Circumference & 0.577 \\
Narrow Waist & 0.552 & Waist Circumference & 0.742 & Horizontal Waist & 0.566 \\
Waist Circumference & 0.526 & Narrow Waist & 0.741 & Abdomen Circumference & 0.559 \\
Abdomen Circumference & 0.512 & Abdomen Circumference & 0.741 & Seat Circumference & 0.510 \\
Seat Circumference & 0.444 & Bicep Circumference Left & 0.720 & Hip Circumference & 0.505 \\
Hip Circumference & 0.442 & Chest & 0.697 & Thigh Circumference Left & 0.479 \\
Bicep Circumference Left & 0.438 & Upper Arm Circumference Right & 0.691 & Thigh Circumference Right & 0.475 \\
Torso Volume & 0.436 & Upper Arm Circumference Left & 0.690 & MidThigh Circumference Left & 0.459 \\
Chest & 0.416 & Bicep Circumference Right & 0.688 & MidThigh Circumference Right & 0.448 \\
Upper Arm Circumference Left & 0.409 & Torso Volume & 0.665 & Narrow Waist & 0.442 \\
\bottomrule
\end{tabular}%
}
\caption{Top 10 most correlated biomarkers with BFP with their Corresponding Correlation Coefficient}
\label{appendix:top-10-bio-BFP}
\end{table}

\begin{table}[ht]

\centering
\resizebox{\textwidth}{!}{%
\renewcommand{\arraystretch}{1.0}
\begin{tabular}{lclclc}
\toprule
\textbf{Male} & \textbf{Corr. Coeff.} & \textbf{Female} & \textbf{Corr. Coeff.} & \textbf{Combined} & \textbf{Corr. Coeff.} \\
\midrule
Surface Area Arm Right & 0.862 & Height (cm) & 0.733 & Surface Area Total & 0.806 \\
Surface Area Arm Left & 0.861 & Surface Area Total & 0.711 & Height (cm) & 0.804 \\
Surface Area Total & 0.851 & Surface Area Leg Left & 0.709 & Surface Area Arm Left & 0.795 \\
Arm Volume Right & 0.834 & Surface Area Leg Right & 0.696 & Surface Area Arm Right & 0.794 \\
Height (cm) & 0.834 & Outside Leg Length Left & 0.675 & Surface Area Leg Left & 0.782 \\
Forearm Circumference Right & 0.832 & Outside Leg Length Right & 0.662 & Surface Area Leg Right & 0.773 \\
Arm Volume Left & 0.831 & Arm Length Left & 0.659 & Surface Area Torso & 0.767 \\
Surface Area Leg Left & 0.829 & Surface Area Torso & 0.658 & Arm Volume Right & 0.765 \\
Surface Area Leg Right & 0.821 & Surface Area Arm Left & 0.654 & Arm Volume Left & 0.758 \\
Surface Area Torso & 0.816 & Hip Circumference & 0.652 & Arm Length Left & 0.749 \\
\bottomrule
\end{tabular}%
}
\caption{Top 10 most correlated biomarkers with BMD with their Corresponding Correlation Coefficient}
\label{appendix:top-10-bio-BMD}
\end{table}


\begin{table}[h!]
\centering
\resizebox{\textwidth}{!}{
\renewcommand{\arraystretch}{1.0}
\setlength{\tabcolsep}{12pt}
\begin{tabular}{l|c|c|c|c|c|c|c|c}
\toprule
\multicolumn{9}{c}{\textbf{ p-Laplacian-1 (ALM)}} \\ \midrule
 \textbf{Dataset}& \textbf{Par./RMSE}  & \textbf{80\%} & \textbf{50\%} & \textbf{33\%} & \textbf{25\%} & \textbf{20\%} & \textbf{10\%} & \textbf{5\%} \\ \hline
\multirow{3}{*}{Male}
  & $p$   & 2.0 & 5.5 & 8.5 & 10.0 & 10.0 & 10.0 & 10.0 \\ 
  & $k$   & 10  & 45  & 55  & 50   & 50   & 50   & 50   \\ \cline{2-9}
  & RMSE  & 8.51 & 8.99 & 9.27 & 9.50 & 9.74 & 10.83 & 12.66 \\ 
  \midrule

\multirow{3}{*}{Female}
  & $p$   & 3.5 & 4.5 & 5.0 & 5.5  & 6.5  & 7.5  & 9.5  \\ 
  & $k$   & 35  & 55  & 60  & 55   & 55   & 50   & 60   \\ \cline{2-9}
  & RMSE  & 10.30 & 10.72 & 11.10 & 11.25 & 11.57 & 12.65 & 14.95 \\ \midrule

\multirow{3}{*}{Combined}
  & $p$   & 3.0 & 2.5 & 2.5 & 3.0  & 3.5  & 10.0 & 10.0 \\ 
  & $k$   & 15  & 10  & 10  & 10   & 10   & 20   & 35   \\ \cline{2-9}
  & RMSE  & 9.97 & 10.46 & 11.05 & 11.42 & 11.84 & 13.21 & 15.20 \\ \bottomrule
\end{tabular}
}
\caption{Optimal Parameters and Corresponding RMSEs of $p$-Laplacian-1  for ALM }
\label{appendix:p-lap-1-ALM}
\end{table}

\begin{table}[h!]
\centering
\resizebox{\textwidth}{!}{
\renewcommand{\arraystretch}{1.0}
\setlength{\tabcolsep}{12pt}
\begin{tabular}{l|c|c|c|c|c|c|c|c}
\toprule
\multicolumn{9}{c}{\textbf{ p-Laplacian-1 (BFP)}} \\ \midrule
 \textbf{Dataset}& \textbf{Par./RMSE}  & \textbf{80\%} & \textbf{50\%} & \textbf{33\%} & \textbf{25\%} & \textbf{20\%} & \textbf{10\%} & \textbf{5\%} \\ \hline
\multirow{3}{*}{Male}
  & $p$   & 3.0 & 2.5 & 3.0 & 3.5 & 10.0 & 10.0 & 10.0 \\ 
  & $k$   & 15  & 10  & 10  & 10  & 45   & 60   & 60   \\ \cline{2-9}
  & RMSE  & 16.61 & 17.68 & 18.90 & 19.80 & 20.51 & 23.40 & 25.56 \\ 
  \midrule

\multirow{3}{*}{Female}
  & $p$   & 3.0 & 3.0 & 4.0 & 4.5  & 5.0  & 6.0  & 8.0  \\ 
  & $k$   & 25  & 35  & 45  & 60   & 55   & 60   & 60   \\ \cline{2-9}
  & RMSE  & 13.11 & 13.78 & 14.22 & 14.51 & 14.78 & 15.57 & 16.62 \\ \midrule

\multirow{3}{*}{Combined}
  & $p$   & 3.5 & 3.5 & 5.0 & 6.0  & 8.0  & 10.0 & 10.0 \\ 
  & $k$   & 15  & 15  & 15  & 15   & 15   & 15   & 15   \\ \cline{2-9}
  & RMSE  & 14.62 & 15.50 & 16.46 & 17.10 & 17.68 & 19.92 & 22.19 \\ \bottomrule
\end{tabular}
}
\caption{Optimal Parameters and Corresponding RMSEs of $p$-Laplacian-1  for BFP }
\label{appendix:p-lap-1-BFP}
\end{table}

\begin{table}[h!]
\centering
\resizebox{\textwidth}{!}{
\renewcommand{\arraystretch}{1.0}
\setlength{\tabcolsep}{12pt}
\begin{tabular}{l|c|c|c|c|c|c|c|c}
\toprule
\multicolumn{9}{c}{\textbf{ p-Laplacian-1 (BMD)}} \\ \midrule
 \textbf{Dataset} & \textbf{Par./RMSE}  & \textbf{80\%} & \textbf{50\%} & \textbf{33\%} & \textbf{25\%} & \textbf{20\%} & \textbf{10\%} & \textbf{5\%} \\ \hline
\multirow{3}{*}{Male}
  & $p$   & 2.5 & 3.0 & 3.0 & 3.0 & 3.5 & 4.5 & 9.5 \\ 
  & $k$   & 20  & 20  & 25  & 20  & 25  & 45  & 60  \\ \cline{2-9}
  & RMSE  & 8.11 & 8.52 & 8.71 & 9.03 & 9.11 & 9.81 & 10.68 \\ 
  \midrule

\multirow{3}{*}{Female}
  & $p$   & 2.0 & 2.0 & 2.0 & 2.5 & 2.5 & 2.5 & 7.0  \\ 
  & $k$   & 10  & 10  & 10  & 25  & 30  & 25  & 60   \\ \cline{2-9}
  & RMSE  & 9.02 & 9.30 & 9.46 & 9.65 & 9.79 & 10.13 & 11.00 \\ 
  \midrule

\multirow{3}{*}{Combined}
  & $p$   & 2.0 & 2.0 & 3.5 & 3.5 & 3.5 & 4.5 & 6.5 \\ 
  & $k$   & 15  & 15  & 40  & 40  & 45  & 50  & 55  \\ \cline{2-9}
  & RMSE  & 8.57 & 8.86 & 9.06 & 9.19 & 9.34 & 9.65 & 10.23 \\ 
\bottomrule
\end{tabular}
}
\caption{Optimal Parameters and Corresponding RMSEs of $p$-Laplacian-1  for BMD }
\label{appendix:p-lap-1-BMD}
\end{table}

\newpage
\begin{table}[h!]
\centering
\resizebox{\textwidth}{!}{
\renewcommand{\arraystretch}{1.0}
\setlength{\tabcolsep}{12pt}
\begin{tabular}{l|c|c|c|c|c|c|c|c}
\toprule
\multicolumn{9}{c}{\textbf{ p-Laplacian-2 (ALM)}} \\ \midrule
\textbf{Dataset} & \textbf{Par./RMSE} & \textbf{80\%} & \textbf{50\%} & \textbf{33\%} & \textbf{25\%} & \textbf{20\%} & \textbf{10\%} & \textbf{5\%} \\ \hline
\multirow{3}{*}{Male}
  & $p$   & 2.0 & 2.0 & 2.5 & 3.0 & 3.0 & 3.0 & 3.5 \\
  & $k$   & 20  & 20  & 30  & 30  & 30  & 30  & 20  \\ \cline{2-9}
  & RMSE  & 8.41 & 8.78 & 9.01 & 9.22 & 9.40 & 10.26 & 11.48 \\ \midrule

\multirow{3}{*}{Female}
  & $p$   & 2.5 & 2.5 & 2.5 & 2.5 & 2.5 & 3.0 & 4.0 \\
  & $k$   & 60  & 60  & 60  & 60  & 40  & 60  & 60  \\ \cline{2-9}
  & RMSE  & 10.57 & 10.87 & 11.07 & 11.20 & 11.41 & 12.13 & 13.83 \\ \midrule

\multirow{3}{*}{Combined}
  & $p$   & 2.0 & 2.5 & 2.5 & 3.0 & 3.0 & 3.5 & 3.0 \\
  & $k$   & 20  & 25  & 25  & 30  & 30  & 60  & 40  \\ \cline{2-9}
  & RMSE  & 11.69 & 12.11 & 12.56 & 12.79 & 13.02 & 13.64 & 14.72 \\
\bottomrule
\end{tabular}
}
\caption{Optimal Parameters and Corresponding RMSEs of $p$-Laplacian-2  for ALM }
\label{appendix:p-lap-2-ALM}
\end{table}

\begin{table}[h!]
\centering
\resizebox{\textwidth}{!}{
\renewcommand{\arraystretch}{1.0}
\setlength{\tabcolsep}{12pt}
\begin{tabular}{l|c|c|c|c|c|c|c|c}
\toprule
\multicolumn{9}{c}{\textbf{ p-Laplacian-2 (BFP)}} \\ \midrule
\textbf{Dataset} &\textbf{Par./RMSE}  & \textbf{80\%} & \textbf{50\%} & \textbf{33\%} & \textbf{25\%} & \textbf{20\%} & \textbf{10\%} & \textbf{5\%} \\ \hline
\multirow{3}{*}{Male}
  & $p$   & 2.0 & 2.0 & 2.5 & 2.5 & 4.5 & 2.0 & 2.0 \\
  & $k$   & 10  & 10  & 10  & 10  & 15  & 15  & 50  \\ \cline{2-9}
  & RMSE  & 18.79 & 20.15 & 20.83 & 22.25 & 22.74 & 25.32 & 26.60 \\ \midrule

\multirow{3}{*}{Female}
  & $p$   & 2.5 & 2.5 & 2.5 & 2.0 & 2.5 & 2.0 & 2.0 \\
  & $k$   & 35  & 40  & 50  & 25  & 60  & 45  & 45  \\ \cline{2-9}
  & RMSE  & 12.99 & 13.27 & 13.54 & 13.72 & 13.99 & 14.48 & 15.07 \\ \midrule

\multirow{3}{*}{Combined}
  & $p$   & 2.0 & 2.0 & 2.5 & 2.5 & 3.0 & 2.5 & 2.0 \\
  & $k$   & 15  & 15  & 20  & 25  & 50  & 60  & 60  \\ \cline{2-9}
  & RMSE  & 20.94 & 21.59 & 22.54 & 22.92 & 23.30 & 24.43 & 25.10 \\
\bottomrule
\end{tabular}
}
\caption{Optimal Parameters and Corresponding RMSEs of $p$-Laplacian-2  for BFP }
\label{appendix:p-lap-2-BFP}

\end{table}

\begin{table}[h!]
\centering
\resizebox{\textwidth}{!}{
\renewcommand{\arraystretch}{1.0}
\setlength{\tabcolsep}{12pt}
\begin{tabular}{l|c|c|c|c|c|c|c|c}
\toprule
\multicolumn{9}{c}{\textbf{ p-Laplacian-2 (BMD)}} \\ \midrule
\textbf{Dataset} & \textbf{Par./RMSE} & \textbf{80\%} & \textbf{50\%} & \textbf{33\%} & \textbf{25\%} & \textbf{20\%} & \textbf{10\%} & \textbf{5\%} \\ \hline
\multirow{3}{*}{Male}
  & $p$   & 2.0 & 2.0 & 2.0 & 2.0 & 2.0 & 2.0 & 2.5 \\
  & $k$   & 50  & 50  & 50  & 30  & 35  & 30  & 55  \\ \cline{2-9}
  & RMSE  & 8.66 & 8.82 & 8.90 & 9.08 & 9.17 & 9.61 & 10.41 \\ \midrule

\multirow{3}{*}{Female}
  & $p$   & 2.5 & 2.0 & 2.0 & 2.0 & 2.0 & 2.0 & 2.5 \\
  & $k$   & 35  & 50  & 35  & 35  & 35  & 30  & 55  \\ \cline{2-9}
  & RMSE  & 9.55 & 9.69 & 9.81 & 9.89 & 9.98 & 10.19 & 10.96 \\ \midrule

\multirow{3}{*}{Combined}
  & $p$   & 2.0 & 2.0 & 2.0 & 2.0 & 2.0 & 2.0 & 2.5 \\
  & $k$   & 55  & 50  & 45  & 45  & 45  & 45  & 55  \\ \cline{2-9}
  & RMSE  & 9.11 & 9.22 & 9.34 & 9.39 & 9.47 & 9.63 & 10.09 \\
\bottomrule
\end{tabular}
}
\caption{Optimal Parameters and Corresponding RMSEs of $p$-Laplacian-2  for BMD }
\label{appendix:p-lap-2-BMD}
\end{table}

\newpage
\begin{table}[h!]
    \centering
    \begin{tabular}{>{\raggedright\arraybackslash}p{2cm}|p{2cm}|c|c|c}
    \toprule
    \multicolumn{5}{c}{\textbf{ Supervised Algorithms (ALM)}} \\ \midrule
\textbf{Model}& \textbf{Par./RMSE} 
                                    & \textbf{Male}
                                    & \textbf{Female}
                                    & \textbf{Combined} \\
    \midrule
    \multirow{2}{*}{Regression}     & Par.               & Traditional      & Ridge           & Bayesian           \\ 
                                    & RMSE               & 8.00            & 9.10            & 9.00            \\
    \midrule
    \multirow{2}{*}{LSSVR}          & Par.               & $\gamma$: 0.001, $C$: 500 & $\gamma$: 0.001, $C$: 1000 & $\gamma$: 0.001, $C$: $1000$\\ 
                                    & RMSE               & 7.47            & 6.03            & 8.46            \\
    \midrule
    \multirow{2}{*}{NN}             & Par.               & Epochs: 300     & Epochs: 300     & Epochs: 200     \\ 
                                    & RMSE               & 7.20            & 9.85            & 8.70            \\
    \midrule
    \multirow{2}{*}{RF}             & Par.               & $n$: 50, $d$: 15 & $n$: 50, $d$: 15 & $n$: 45, $d$: 25 \\ 
                                    & RMSE               & 7.96            & 10.52           & 10.06           \\
    \midrule
    \multirow{2}{*}{SVR}            & Par.               & $\epsilon$: 0.7, $C$: 50  & $\epsilon$: 0.3, $C$: 1.0  & $\epsilon$: 0.4, $C$: 10  \\ 
                                    & RMSE               & 6.30            & 8.97            & 7.83            \\
    \midrule
    \multirow{2}{*}{XGBoost}        & Par.               & $n$: 35, $d$: 05 & $n$: 35, $d$: 05 & $n$: 35, $d$: 05 \\ 
                                    & RMSE               & 8.36            & 10.82           & 10.27           \\
    \bottomrule
    \end{tabular}
      \caption{Optimal Parameters and Corresponding RMSEs of Supervised Algorithms for ALM}
    \label{appendix:supervised-ALM_performance}
\end{table}

\begin{table}[h!]
    \centering
    \begin{tabular}{>{\raggedright\arraybackslash}p{2cm}|p{2cm}|c|c|c}
    \toprule
      \multicolumn{5}{c}{\textbf{ Supervised Algorithms (BFP)}} \\ \midrule
    \textbf{Model} & \textbf{Par./RMSE} 
                                    & \textbf{Male}
                                    & \textbf{Female}
                                    &\textbf{Combined} \\
                            
    \midrule
    \multirow{2}{*}{Regression}     & Par.               & Ridge         & Bayesian           & Bayesian           \\ 
                                    & RMSE               & 18.00         & 14.60           & 17.00           \\
    \midrule
    \multirow{2}{*}{LSSVR}          & Par.               & $\gamma$: 0.01, $C$: 25 & $\gamma$: 0.01, $C$: 25 & $\gamma$: 0.001, $C$: 250 \\ 
                                    & RMSE               & 11.47         & 12.09           & 10.99           \\
    \midrule
    \multirow{2}{*}{NN}             & Par.               & Epochs: 300   & Epochs: 400     & Epochs: 250     \\ 
                                    & RMSE               & 14.20         & 12.00           & 12.80           \\
    \midrule
    \multirow{2}{*}{RF}             & Par.               & $n$: 50, $d$: 15 & $n$: 50, $d$: 20 & $n$: 50, $d$: 15 \\ 
                                    & RMSE               & 15.92         & 12.44           & 14.13           \\
    \midrule
    \multirow{2}{*}{SVR}            & Par.               & $\epsilon$: 0.5, $C$:0.7 & $\epsilon$: 1.0, $C$: 0.1 & $\epsilon$: 1.0, $C$: 1.0 \\ 
                                    & RMSE               & 14.92         & 11.72           & 13.28           \\
    \midrule
    \multirow{2}{*}{XGBoost}        & Par.               & $n$: 35, $d$: 05 & $n$: 30, $d$: 05 & $n$: 35, $d$: 05 \\ 
                                    & RMSE               & 16.40         & 12.89           & 14.55           \\
    \bottomrule
    \end{tabular}

    \caption{Optimal Parameters and Corresponding RMSEs of Supervised Algorithms for BFP}
    \label{appendix:supervised-bfp_performance}
\end{table}

\begin{table}[]
    \centering
    \begin{tabular}{>{\raggedright\arraybackslash}p{2cm}|p{2cm}|c|c|c}
    \toprule
      \multicolumn{5}{c}{\textbf{ Supervised Algorithms (BMD)}} \\ \midrule
  \textbf{Model} &\textbf{Par./RMSE} 
                                    & \textbf{Male}
                                    & \textbf{Female}
                                    &\textbf{Combined} \\
    \midrule
    \multirow{2}{*}{Regression}     & Par.               & Ridge         & Bayesian           & Ridge           \\ 
                                    & RMSE               & 8.10          & 9.25            & 8.00            \\
    \midrule
    \multirow{2}{*}{LSSVR}          & Par.               & $\gamma$: 0.001, $C$: 250 & $\gamma$: 0.001, $C$: 100 & $\gamma$: 0.01, $C$: 10 \\ 
                                    & RMSE               & 7.59          & 7.05            & 8.09            \\
    \midrule
    \multirow{2}{*}{NN}             & Par.               & Epochs: 200   & Epochs: 350     & Epochs: 400     \\ 
                                    & RMSE               & 7.35          & 8.60            & 7.90            \\
    \midrule
    \multirow{2}{*}{RF}             & Par.               & $n$: 45, $d$: 15 & $n$: 50, $d$: 20 & $n$: 50, $d$: 10 \\ 
                                    & RMSE               & 7.72          & 8.09            & 7.94            \\
    \midrule
    \multirow{2}{*}{SVR}            & Par.               & $\epsilon$: 0.05, $C$: 0.2 & $\epsilon$: 0.05, $C$: 0.05 & $\epsilon$: 0.05, $C$: 0.4 \\ 
                                    & RMSE               & 6.97          & 8.36            & 7.48            \\
    \midrule
    \multirow{2}{*}{XGBoost}        & Par.               & $n$: 20, $d$: 05 & $n$: 15, $d$: 05 & $n$: 20, $d$: 05 \\ 
                                    & RMSE               & 8.22          & 8.37            & 8.31            \\
    \bottomrule
    \end{tabular}
    \caption{Optimal Parameters and Corresponding parameters of supervised algorithms for BMD}
    \label{appendix:supervised-bmd_performance}
\end{table}

\end{document}